\documentclass{article} % For LaTeX2e
\usepackage{iclr2025_conference,times}

\usepackage{amsmath,amsfonts,bm}

\def\eqref#1{equation~\ref{#1}}
\def\1{\bm{1}}

\DeclareMathAlphabet{\mathsfit}{\encodingdefault}{\sfdefault}{m}{sl}
\SetMathAlphabet{\mathsfit}{bold}{\encodingdefault}{\sfdefault}{bx}{n}

\usepackage{url}

\usepackage{booktabs}       % professional-quality tables
\usepackage{amsfonts}       % blackboard math symbols
\usepackage{nicefrac}       % compact symbols for 1/2, etc.
\usepackage{microtype}      % microtypography
\usepackage{xcolor}         % colors
\usepackage{amsmath}
\usepackage{graphicx}
\usepackage{enumitem}
\usepackage{bbm}
\usepackage{subcaption}
\usepackage{multirow}
\usepackage{fancyvrb}
\usepackage{wrapfig}
\usepackage[hidelinks]{hyperref}
\usepackage{mdframed}

\newcommand{\fff}[0]{\text{F$^3$}}
\newcommand{\fest}[0]{\text{F$^3$ED}}
\newcommand{\dataset}[0]{\text{F$^3$Set}}

\title{\dataset{}: Towards Analyzing Fast, Frequent, and Fine-grained Events 
from Videos}

\author{
    Zhaoyu Liu\textsuperscript{1,2}, 
    Kan Jiang\textsuperscript{2}, 
    Murong Ma\textsuperscript{2}, 
    Zhe Hou\textsuperscript{3},
    Yun Lin\textsuperscript{4},
    Jin Song Dong\textsuperscript{2}\\
    \textsuperscript{1}Ningbo University \textsuperscript{2} National University of Singapore \textsuperscript{3} Griffith University \\
    \textsuperscript{4} Shanghai Jiao Tong University \\
    \texttt{\{liuzy, jiangkan\}@nus.edu.sg, murongma@u.nus.edu} \\ 
    \texttt{z.hou@griffith.edu.au, lin\_yun@sjtu.edu.cn, dcsdjs@nus.edu.sg}
}

\iclrfinalcopy
\begin{document}

\maketitle

\vspace{-10px}
\begin{abstract}
% Analyzing Fast, Frequent, and Fine-grained (\fff{}) events presents a significant challenge in video analytics and multi-modal LLMs. Although current methods exhibit efficacy on public benchmarks, they struggle to identify events that satisfy all the \fff{} criteria with high accuracy due to challenges such as motion blur and subtle visual discrepancies. To address this, we introduce \dataset{}, a new benchmark dataset built on tennis video specifically for \fff{} event detection. \dataset{} is characterized by its extensive scale and comprehensive detail, encompassing over 1,000 event types with precise timestamps and supporting multi-level granularity. We evaluated popular temporal action understanding methods on \dataset{}, revealing substantial challenges for existing techniques. Additionally, we propose a new method, \fest{}, for \fff{} event detections, achieving superior performance. 
% % Using tennis as a case study, we demonstrate the utility of \fff{} sequences for advanced automated strategic analytics. 
% The dataset, model and the benchmark code are available \href{https://github.com/F3Set/F3Set}{online}.
\vspace{-5px}
Analyzing Fast, Frequent, and Fine-grained (\fff{}) events presents a significant challenge in video analytics and multi-modal LLMs. Current methods struggle to identify events that satisfy all the \fff{} criteria with high accuracy due to challenges such as motion blur and subtle visual discrepancies. 
To advance research in video understanding, 
we introduce \dataset{}, a benchmark 
that consists of video datasets
%built primarily on tennis video 
for precise \fff{} event detection. 
Datasets in \dataset{} are characterized by their extensive scale and comprehensive detail, usually encompassing over 1,000 event types with precise timestamps and supporting multi-level granularity. Currently \dataset{} contains several sports datasets, and this framework may be extended to other applications as well. We evaluated popular temporal action understanding methods on \dataset{}, revealing substantial challenges for existing techniques. Additionally, we propose a new method, \fest{}, for \fff{} event detections, achieving superior performance. 
The dataset, model, and benchmark code are available at \url{https://github.com/F3Set/F3Set}.
\end{abstract}
\vspace{-10px}

\section{Introduction}
\vspace{-5px}

Recognizing sequences of fast (fast-paced), frequent (many actions in a short period), and fine-grained (diverse types) events with precise timestamps (with a tolerance of 1-2 frames) is a challenging problem for both current video analytics methods and multi-modal large language models (LLMs). Despite advances in fine-grained action recognition [\citenum{lea2016learning,shao2020finegym,munro2020multi}], temporal action localization [\citenum{shou2016temporal,chao2018rethinking,liu2022empirical,shi2023tridet}], segmentation [\citenum{wang2020boundary,li2021temporal,yi2021asformer,behrmann2022unified}], and video captioning [\citenum{wang2018reconstruction,pei2019memory,luo2020univl,lin2022swinbert}], limited focus has been focused on this problem. This task is critical for various real-world applications, such as sports analytics, where action forecasting [\citenum{felsen2017will,wang2022shuttlenet}], strategic and tactical analysis [\citenum{liu2023insight,liu2024exploring,liu2024strategy,liu2025analyzing}], and player performance evaluation [\citenum{decroos2019actions,pappalardo2019playerank}] depend on understanding \emph{detailed} of event sequences. Other examples include industrial inspection [\citenum{liu2022videopipe}], crucial for detecting subtle irregularities in high-speed production lines to ensure quality and safety; computer vision in autonomous driving [\citenum{huang2018apolloscape}], essential for accurate and instantaneous vehicle control and obstacle detection; and surveillance [\citenum{oh2011large}], important for the precise identification of abnormal or sudden events to enhance security. However, existing methods and datasets foundational to their development only \emph{partially} address the \fff{} scenario.

To facilitate the study of \fff{} events understanding, we propose a new benchmark, \dataset{}, for precise temporal events detection and recognition. \dataset{} datasets usually have a large number of event types (on the order of 1,000), annotated with exact timestamps, and offer multi-level granularity to capture comprehensive event details. Although \fff{} is a general problem, creating such a dataset requires domain-specific knowledge for labeling and processing; thus, in this paper, we use tennis as a case study. We also introduce a general annotation pipeline and toolchain to support domain experts in creating new \fff{} datasets. Using this pipeline, we have also been building datasets for table tennis and badminton, and a community of users is actively expanding these with other applications.

\begin{figure}
\centering
    \includegraphics[width=0.9\linewidth]{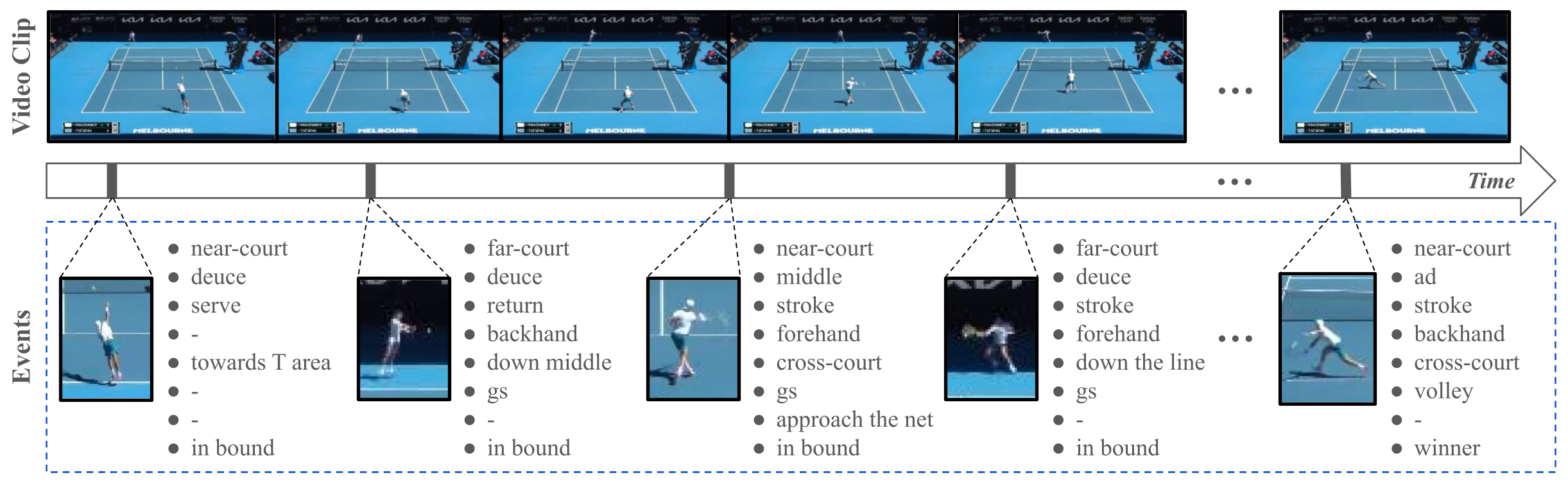}
    \vspace{-5px}
\caption{Example of detecting fast, frequent, and fine-grained events with precise moments.}
\vspace{-10px}
\label{fig:vid2seq}
\end{figure}

Unlike other video analysis tasks, tennis actions are characterized by their rapid succession and diversity, as illustrated in Figure~\ref{fig:vid2seq}. Understanding detailed event attributes such as shot direction, technique, and outcome is essential. For example, analyzing patterns in serve directions (e.g., ``T'', ``body'', ``wide'', defined in Appendix~\ref{lex}) or success rates can reveal players' habits and skill levels, offering strategic insights for competitive advantage [\citenum{dong2015sports}]. This detailed analysis supports coaches and players in developing tailored strategies against different opponents [\citenum{dong2023sports,liu2024pcsp}]. However, detecting \fff{} events from videos poses significant challenges, such as subtle visual differences, motion-induced blurring, and the need for precise event localization. Current video understanding methods are inadequately equipped to address these challenges. For instance, traditional fine-grained action recognition [\citenum{damen2018scaling,shao2020finegym,jiang2020deep}] assigns a single label to an entire video rather than identifying a sequence of events. Temporal action localization (TAL) and temporal action segmentation (TAS) often depend on pre-trained or modestly fine-tuned input features [\citenum{liu2023survey,ding2023temporal}], which lack the specificity required to capture the subtle and domain-specific visual details necessary for recognizing diverse events with temporal precision. Some studies [\citenum{hong2022spotting,lin2022swinbert,liu2023recognizing}] attempt to address these issues through \emph{dense} frame sampling and end-to-end training. However, this makes targeted events temporally sparse (e.g., only a few events over hundreds of consecutive frames). As a result, long-term temporal correlation modules on dense visual features struggle to capture event-wise causal correlations effectively.

Moreover, Large Language Models (LLMs) [\citenum{openai2023gpt4,team2023gemini,liu2024visual}] have expanded their capabilities to include multi-modal inference, encompassing text, visuals, and audio. Recognizing the potential, we conducted preliminary experiments on \dataset{} using GPT-4 and observed that it understood basic video contexts, such as sports types, contextual information (e.g., court type and scoreboard), and simple actions. However, it struggles with understanding \fff{} events and temporal relations between frames (e.g., shot directions). See Appendix~\ref{gpt} for details. Consequently, GPT-4 yields poor results compared to the other methods for \fff{} problems, and we do not use it in the experiment. 
% Furthermore, realizing the shortcoming of current LLMs on identifying \fff{} events, our benchmark can contribute towards a ``VideoNet'' (like ImageNet [\citenum{deng2009imagenet}]) for future LLM's benchmarking and fine-tuning.
{\color{black} By introducing \dataset{}, we hope it can help advance multi-modal LLM capabilities in \fff{} video understanding in the future.}

Leveraging \dataset{}, we extensively evaluate existing temporal action understanding methods, aiming to reveal the challenges of \fff{} event understanding. 
To provide guidelines for future research, we conduct a number of ablation studies on modeling choices. {\color{black}Addressing the shortcomings of existing methods, we also propose a simple yet efficient model, \fest{}, that is designed for \fff{} event detection tasks and can be trained quickly on a single GPU.} It outperforms existing models and can serve as a baseline for further development.

% contributions
% \vspace{5px}\noindent\textbf{Contributions.} 
\vspace{-5px}\paragraph{Contributions.}
The key contributions of this paper are as follows:
\vspace{-5px}
% \begin{itemize}[noitemsep,leftmargin=15pt]
% \item The creation of \dataset{}, a new benchmark dataset that features a rich collection of over 1,000 precisely timestamped event types, designed to challenge and advance the state-of-the-art in temporal action understanding.
% \item Comprehensive evaluation of leading temporal action understanding methods on \dataset{}, through rigorous testing and ablation studies, establishing a benchmark for future research.
% \item Demonstration of the practical applications of \dataset{} in real-world scenarios, using tennis as a case study to highlight the dataset’s utility in enhancing fine-grained event analysis.
% \end{itemize}
\begin{itemize}[noitemsep,leftmargin=15pt]
    \item We create \dataset{}, a new benchmark with datasets that feature over 1,000 precisely timestamped event types with multi-level granularity, designed to challenge and advance the state-of-the-art in temporal action understanding.
    \item We introduce a general annotation toolchain that enables domain experts to create new \fff{} datasets.
    \item We propose an end-to-end model named \fest{}, which can accurately detect \fff{} event sequences from videos through visual features and contextual sequence refinement {\color{black}on a single GPU}.
    \item We assess the performance of leading temporal action understanding methods on \dataset{} through comprehensive evaluations and ablation studies and analyze the results.
    % \item  We demonstrate potential applications of analyzing \fff{} events using tennis as a case study.
\end{itemize}
\vspace{-10px}

% We make the following contributions: 
% \begin{itemize}[noitemsep,leftmargin=15pt]
%     \item We construct \dataset{}, a new large-scale video dataset for \fff{} event detection that features multi-level granularity and accurately captured timestamp for each hitting moment.
%     \item We develop a benchmark for \dataset{} and assess the performance of leading temporal action understanding methods on \dataset{} through comprehensive evaluations and ablation studies.
%     \item  We demonstrate potential applications of analyzing \fff{} events using tennis as a case study.
% \end{itemize}

% \vspace{5px}\noindent\textbf{Contributions.} We make the following contributions: 
% \begin{enumerate}
% % method
%     \item We propose an end-to-end \fest{}, which can accurately locate and recognize complex event sequences from videos through visual features and contextual sequence refinement.
%     % dataset
%     \item We build a new sports video dataset \dataset{}, which is the first large-scale tennis video dataset with multi-level event granularity classes and precise hitting moments.
%     % performance
%     \item We benchmark \fest{} with recent TAL and TAS methods and conduct case studies to demostrate the effectiveness and usefulness of our model.
% \end{enumerate}

\section{Related Work}
\vspace{-5px}

\paragraph{Existing \fff{} related datasets.}
Although datasets have been developed for temporal action understanding, few focus on the \fff{} events. Table~\ref{tab:sports_datasets} compares existing datasets with \dataset{} by scale (``\# Vid'', ``\# Clips'') and characteristics like action speed (``Evt. Len.''), frequency (``Evt. / sec''), and granularity (``\# Classes''), {\color{black} which correspond to ``fast'', ``frequent'', and ``fine-grained'' respectively}. Datasets such as THUMOS14 [\citenum{idrees2017thumos}] and Breakfast [\citenum{Kuehne16end}] focus on coarse-grained actions, where background context provides clear cues, and actions span seconds to minutes. In contrast, FineAction [\citenum{liu2022fineaction}] and ActivityNet [\citenum{caba2015activitynet}] cover a wide range of daily activities with diverse action categories, while FineGym [\citenum{shao2020finegym}] delves into detailed action types within gymnastics. Like FineGym, \dataset{} emphasizes domain-specific granularity with subtle visual differences but encounters additional challenges due to faster and more frequent actions. Besides, unlike FineGym’s typical single-player focus, \dataset{} (e.g., tennis) features two players and a fast-moving ball, with both players rapidly moving across the court, occupying only small portions of the scene, thus increasing task difficulty. 
CCTV-Pipe [\citenum{liu2022videopipe}] targets temporal defect detection in urban pipe systems, providing single-frame annotations for rapid event detection, though it is limited in frequency and event types.
% This introduces complexities not found in other fine-grained datasets due to the heightened pace and frequency of actions in \dataset{}.
% Fine-grained datasets, such as FineAction [\citenum{liu2022fineaction}] and ActivityNet [\citenum{caba2015activitynet}], emphasize the breadth and diversity of action categories across various daily activities (e.g., household, sports exercise, personal care), whereas FineGym [\citenum{shao2020finegym}] concentrates on the depth of action types within the domain-specific area of gymnastics. Like FineGym, \dataset{} focuses on depth of granularity, a challenging aspect due to the subtle visual distinctions between actions. However, the actions in FineGym are not as fast-paced and frequent as the ones in \dataset{}. Furthermore, FineGym only contains \textbf{one player} usually clearly shown and located in the \textbf{center} of the scene. However, in FineTennis, \textbf{two players} are located separately on both sides of the court with a small fast moving tennis ball. Both are \textbf{moving rapidly} from one place to another and only occupy \textbf{small proportions} of the entire scene. All these differences increase the difficulty of our task. However, unlike other fine-grained datasets, \dataset{} introduces additional complexities due to its faster and more frequent actions.
Research in the sports domain has explored the detection of fast and frequent actions. FineDiving [\citenum{xu2022finediving}] segments diverse diving events, while 
% Tennis [\citenum{hong2022spotting}], 
ShuttleSet [\citenum{wang2023shuttleset}] and P$^2$ANet [\citenum{bian2024p2anet}] focus on identifying strokes in fast-paced racket sports. {\color{black} Volleyball [\citenum{ibrahim2016hierarchical}] and NSVA (basketball) [\citenum{wu2022sports}] focus on team sports understanding and video captioning, while SoccerNetV2 [\citenum{deliege2021soccernet}] ball action spotting task focus on identifying the timing and type of ball-related actions.}
However, these datasets typically cover coarser event types and are limited to specific \fff{} aspects. 

In contrast, our proposed \dataset{} is characterized by 
% 1) rapidity, with events occurring instantaneously; 2) frequency, with approximately one event per second; and 3) a high degree of granularity, featuring a significantly larger number of detailed event classes. These attributes introduce novel challenges for temporal action understanding methodologies.
1) \emph{rapid} events that occur instantaneously, 2) \emph{high frequency} of approximately one event per second, and 3) \emph{extensive granularity} with a larger number of detailed event classes. These attributes introduce novel challenges.
% for temporal action understanding methodologies.
\vspace{-5px}

{\color{black} \paragraph{\fff{} event understanding}
Detecting \fff{} events poses unique challenges due to their rapid temporal dynamics, high occurrence rates, and subtle visual distinctions, requiring precise temporal and contextual understanding. Fine-grained action detection has been explored in tasks covering diverse daily activities [\citenum{caba2015activitynet, liu2022fineaction}], using features extracted by video encoders pre-trained on datasets like Kinetics-400 [\citenum{kay2017kinetics}] and a detection head for classification. However, such pre-trained extractors often miss domain-specific nuances. Domain-specific methods in FineGym [\citenum{shao2020finegym}] and FineDiving [\citenum{shao2020finegym}] utilize end-to-end training to incorporate domain knowledge. These methods often encode videos into non-overlapping snippets or downsample frames, yielding coarse temporal features insufficient for detecting fast-paced events spanning only 1–2 frames. Related works such as ShuttleSet [\citenum{wang2023shuttleset}] and P$^2$ANet [\citenum{bian2024p2anet}] address fast and frequent event detection in racket sports by employing end-to-end models that extract frame-wise features and use detection heads (e.g., BMN [\citenum{lin2019bmn}] or GRU [\citenum{chung2014empirical}]) to classify each frame. To address class imbalance, the loss weight of the foreground classes is set higher than the background during training [\citenum{hong2022spotting}]. While these approaches achieve precise temporal spotting, their scalability to larger action classes is limited by challenges like long-tail class distributions and inadequate modeling of event-wise correlations. Our proposed \fest{} overcomes these issues through frame-wise dense processing, a multi-label classification head to handle minor event differences and class imbalances, and a contextual module to refine predictions by leveraging event-wise causal relationships, enhancing both precision and robustness in \fff{} event detection.}
\vspace{-10px}

\begin{table}
\small
\caption{Comparison of existing {\color{black} \fff{} related} datasets and $\dataset{}$. ``Evt. Len.'' is the average duration of each event, and ``\# Evt. / sec'' is the average number of events per second.}
\vspace{-1.5mm}
    \centering
    \begin{tabular}{lcccc c cc}
        \toprule
        Datasets  & \# Vid. & \# Clips. & Avg. Clip Len.  & \# Classes & Evt. Len. & \# Evt. / sec %(\# Event / second) 
        \\
        % & \# Event types & \# sub-classes & Avg. Dur. (sec) \\
        % \midrule
        % \multicolumn{5}{l}{\textit{(a) Coarse-grained}} \vspace{1mm} \\
        % THUMOS14 [\citenum{}]  & - & 413  & 264.xs & 20 &4.3s & \\
        % Breakfast [\citenum{}]  & - &   & s & 48 &s & \\
        \midrule
        \multicolumn{5}{l}{\textit{(a) Fine-grained}} \vspace{1mm} \\
        FineAction [\citenum{liu2022fineaction}]  & - & 16,732  & 149.5s & 101 &6.9s &0.3 \\
        ActivityNet [\citenum{caba2015activitynet}] & - & 19,994 & 116.7s & 200 & 49.2s &0.01 \\
        FineGym [\citenum{shao2020finegym}] & 303 & 32,697 & 50.3s & 530 & 1.7s &0.3  \\
        \midrule
        \multicolumn{5}{l}{\textit{(b) Fast}} \vspace{1mm} \\
        CCTV-Pipe [\citenum{liu2022videopipe}]  & 575 & 575 & 549.3s & 16 &{$< 0.1$s} &0.02 \\
        {\color{black}SoccerNetV2 [\citenum{deliege2021soccernet}]} & {\color{black}9} & {\color{black}9} & {\color{black}99.6min} &{\color{black}12} & {\color{black}$< 0.1$s} & {\color{black}0.3}\\
        % TenniSet [\citenum{faulkner2017tenniset}] &5 &746 &13.8s &11 & <1s & 0.8 \\
        % BadmintonDB [\citenum{ban2022badmintondb}] &9 &811 &20 &2 ($sc_1 + sc_6$) & 12.1 \\
        \midrule
        \multicolumn{5}{l}{\textit{(c) Frequent}} \vspace{1mm} \\
        FineDiving [\citenum{xu2022finediving}] & 135 & 3,000 &4.2s & 29 & 1.1s & $\sim$1  \\
        \midrule
        \multicolumn{5}{l}{\textit{(d) Fast \& Frequent}} \vspace{1mm} \\
        % Tennis [\citenum{hong2022spotting}] &28 & 3,791 & 13.8s &6 & $< 0.1$s &$\sim$1 \\
        ShuttleSet [\citenum{wang2023shuttleset}] & 44 & 3,685 &10.9s & 18 & $< 0.1$s & $\sim$1\\
        P$^2$ANet [\citenum{bian2024p2anet}] &200  & 2,721 &360.0s & 14 & $< 0.1$s & $\sim$2 \\
        % BadmintonDB & 811 & 20 &  12.1s & 1 $\sim$ 54 \\
        % MultiSports & General & 37,701 & 66 & 1.0s & 1 \\
        \midrule
        \multicolumn{5}{l}{\textit{(d) Fast \& Frequent \& Fine-grained}} \vspace{1mm} \\
        \textbf{$\dataset{}$} &114 & 11,584 &8.4s & {1,108} & {$< 0.1$s} & {$\sim$1} \\
        \bottomrule
    \end{tabular}
    \vspace{-15px}
    \label{tab:sports_datasets}
\end{table}
\normalsize

% \textbf{Fine-grained video understanding}

% \textbf{Sports video analytics} is a growing field in video analytics. It aims to apply computer vision and deep learning techniques to automatically obtain match insights from different sources of sports recordings. It contains subtasks such as court detection [\citenum{cioppa2022scaling}], player and ball tracking [\citenum{voeikov2020ttnet,huang2019tracknet}], dense video captioning [\citenum{mkhallati2023soccernet}], action recognition and detection [\citenum{li2021multisports,hong2022spotting}]. Furthermore, there are also several public sports-related video datasets. For instance, datasets like FineGym [\citenum{shao2020finegym}] and FineDiving [\citenum{xu2022finediving}] focus on fine-grained action recognition and segmentation, Volleyball [\citenum{ibrahim2016hierarchical}] and NSVA (basketball) [\citenum{wu2022sports}] focus more on group activity understanding and captioning, and SoccerNet [\citenum{deliege2021soccernet}] involves various tasks such as action spotting, camera calibration, player re-identification and tracking. However, few works focus on recognizing \fff{} event sequences in racket sports videos.
{\color{black} \section{{\dataset{}: A Benchmark Dataset for \fff{} Event Detection}}} \label{sec:dataset}
\vspace{-5px}
Recognizing the limitations in
% scalability and granularity of 
existing video datasets for \fff{} event understanding, we introduce $\dataset{}$, 
% a new fine-grained video dataset, 
a new benchmark for precise temporal \fff{} events detection and recognition. 
% containing detailed \fff{} event information and the precise timestamp. 
{\color{black} Given the need for domain-specific expertise in creating \fff{} datasets, this section uses \textbf{tennis} as a \textbf{case study} to illustrate \dataset{}'s event descriptions, construction process, and key properties. We also propose a general annotation pipeline and toolchain that empowers domain experts to develop new \fff{} datasets for diverse applications.}
% \dataset{} consists of multiple domains, including tennis (both singles and doubles matches), badminton, and table tennis. 
% % We are actively expanding its scope and coverage to more domains. 
% The majority of current \dataset{} consists of tennis single matches. Still, we are actively expanding its scope and coverage (see \href{https://github.com/F3Set/F3Set/tree/main/data/}{link}).
% , such as tennis doubles, badminton, and table tennis. 
% In this section, we take \emph{tennis single} as an example to discuss the \fff{} event definition, annotation toolchain, and dataset properties. 
% Using our toolchain, we have built \fff{} datasets for badminton, table tennis, and tennis doubles. A community of users are actively expanding these with other applications (see \href{https://github.com/F3Set/F3Set/tree/main/data/}{link}).
{\color{black} Applying the same approach, we have also built \fff{} datasets for \textbf{other domains}, including tennis doubles, badminton, and table tennis (see \href{https://github.com/F3Set/F3Set/tree/main/data/}{\color{black}{\underline{link}}}).}
% and extended easily generalized to other \fff{} domains.

{\color{black}\subsection{\dataset{} event description}
\vspace{-5px}
We use \textbf{tennis} to illustrate \fff{} event descriptions, introducing key lexicon and defining \fff{} events.  Datasets have been built for \textbf{other \fff{}domains}, including tennis doubles, badminton, and table tennis, with similar event definitions. Details are in Appendix~\ref{other-fff-events}.}
\vspace{-5px}

\paragraph{Lexicon.}
{\color{black} A tennis court is divided into deuce, middle, and ad regions. The initial shot, a ``serve,'' targets the T, Body (B), or Wide (W) areas. A ``return'' follows if the receiver's shot lands in bounds. Subsequent shots, or ``strokes'', can be directed ``cross-court'' (CC), ``down the line'' (DL), ``down the middle'' (DM), ``inside-in'' (II), or ``inside-out'' (IO) using either ``forehand'' (fh) or ``backhand'' (bh). Players may ``approach'' (apr) the net on shorter balls. Shot techniques include ``ground stroke/top spin'' (gs), ``slice'', ``volley'', and ``lob'', with outcomes: ``in-bound'', ``winner'', ``forced error'', or ``unforced error''.}
{\color{black} More detailed definitions can be found in Appendix~\ref{lex}.}
\vspace{-5px}

\begin{figure}
\centering
    \includegraphics[width=0.8\linewidth]{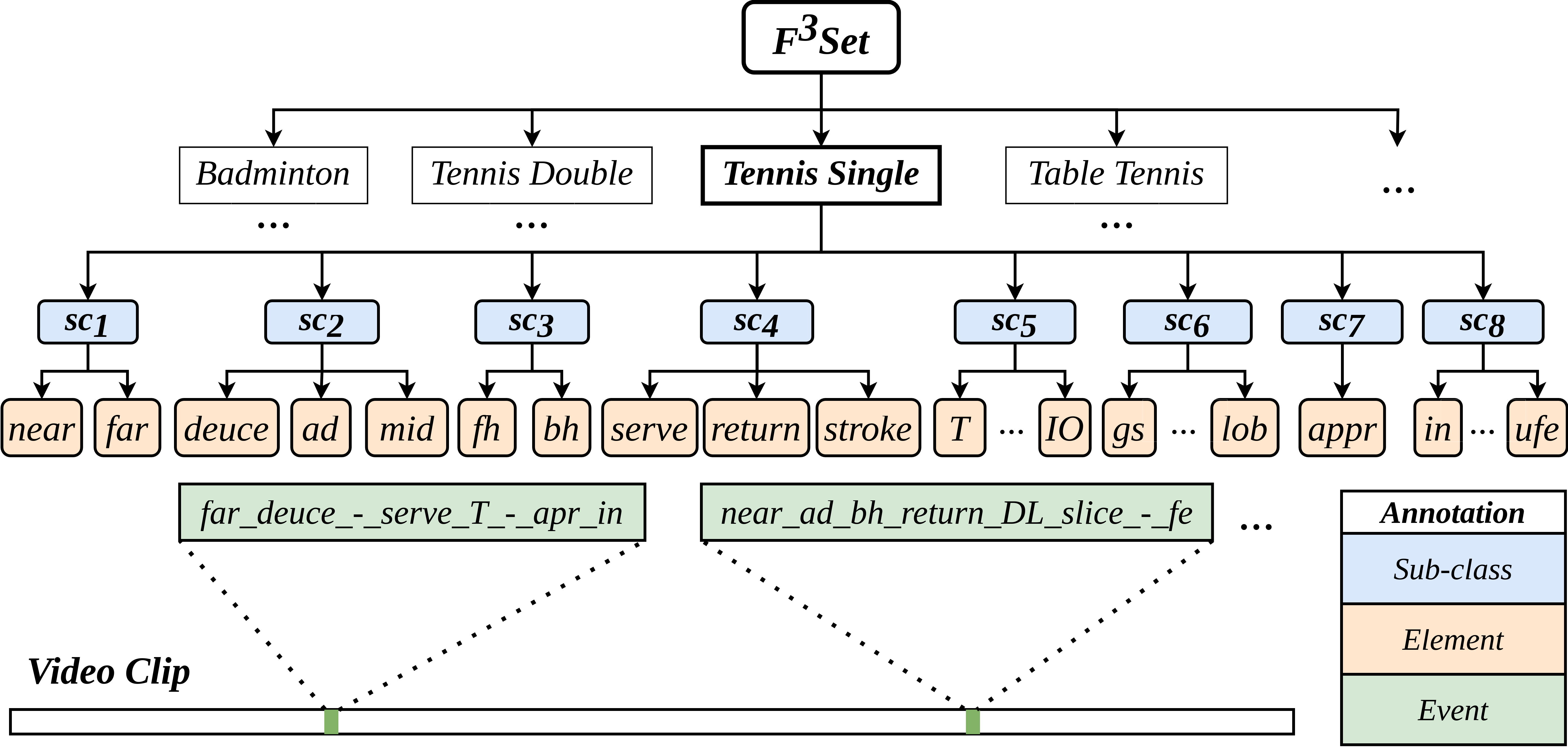}
    \vspace{-5px}
\caption{Breakdown of \dataset{} event class annotation.}
\vspace{-10px}
\label{fig:classes}
\end{figure}

\paragraph{\fff{} events.} 
% A tennis event describes detailed information on each shot. We follow the technical tennis terms based on official terminology from the USTA\footnote{\url{https://www.usta.com/en/home/improve/tips-and-instruction.html}}. 
Formally, each event consists of 8 \emph{sub-classes}, denoted as $sc_1, sc_2, ..., sc_8$: 
\vspace{-5px}
\small
{\color{black}\begin{description}
% \begin{itemize}[noitemsep,leftmargin=15pt]
\item $sc_1$ -- \textit{hit by which player}: (1) near- or (2) far-end player; 
\item $sc_2$ -- \textit{hit from which court location}: (3) deuce, (4) middle, or (5) ad court; 
\item $sc_3$ -- \textit{hit at which side of the body}: (6) forehand or (7) backhand; 
\item $sc_4$ -- \textit{shot type}: (8) serve, (9) return, or (10) stroke; 
\item $sc_5$ -- \textit{shot direction}: (11) T, (12) B, (13) W, (14) CC, (15) DL, (16) DM, (17) II, or (18) IO; 
\item $sc_6$ -- \textit{shot technique}: (19) gs, (20) slice, (21) volley, (22) lob, (23) drop, or (24) smash; 
\item $sc_7$ -- \textit{player movement}: (25) approach; 
\item $sc_8$ -- \textit{shot outcome}: (26) in, (27) winner, (28) forced error, or (29) unforced error.
% \end{itemize}
\end{description}}
\normalsize
\vspace{-5px}
Altogether, there are 29 \emph{elements} and 1,108 \emph{event types} based on various combinations (Figure~\ref{fig:classes}).

{\color{black} Similarly, for other domains, badminton contains 6 \emph{sub-classes}, 28 \emph{elements} and 1008 \emph{event types}; table tennis contains 7 \emph{sub-classes}, 23 \emph{elements} and 1296 \emph{event types}; and tennis doubles contain 26 \emph{elements} and 744 \emph{event types}. Compared to existing racket sports video datasets [\citenum{bian2024p2anet,wang2023shuttleset}], \dataset{} offers additional dimensions, such as shot direction and outcomes, which are crucial for identifying playing patterns and success rates.
Please refer to Appendix~\ref{other-fff-events} for more details.}
\vspace{-5px}

{\color{black}\subsection{\dataset{} Dataset construction} \label{prepare}
\vspace{-3px}
\paragraph{Video collection.}  
For tennis, we collected publicly available high-resolution singles matches (2012–2023) from YouTube, including Grand Slams, Olympics, and major ATP/WTA tournaments. The dataset includes various court surfaces (hard, clay, grass), male and female players, and both right- and left-handed competitors. These videos feature complete rallies, match footage, and detailed player data. Similar criteria were used for tennis doubles, badminton, and table tennis videos.}
% We collected high-resolution videos of singles tennis matches spanning from 2012 to 2023, sourced from YouTube. These matches encompass Grand Slams, Olympic games, and major ATP and WTA tournaments. The videos feature a variety of court surfaces—including hard, clay, and grass—and showcase both male and female players, as well as competitors with different handedness (right-handed and left-handed). 
% % We also collected videos of tennis doubles, badminton, and table tennis matches for both professional and NCAA-Division1 (college) levels. We are actively expanding the scale and coverage of our \dataset{}.
% They provide comprehensive content that includes complete tennis rallies, extensive match footage, and detailed player information.
\vspace{-5px}

\begin{figure}[t!]
\centering
    \includegraphics[width=0.9\linewidth]{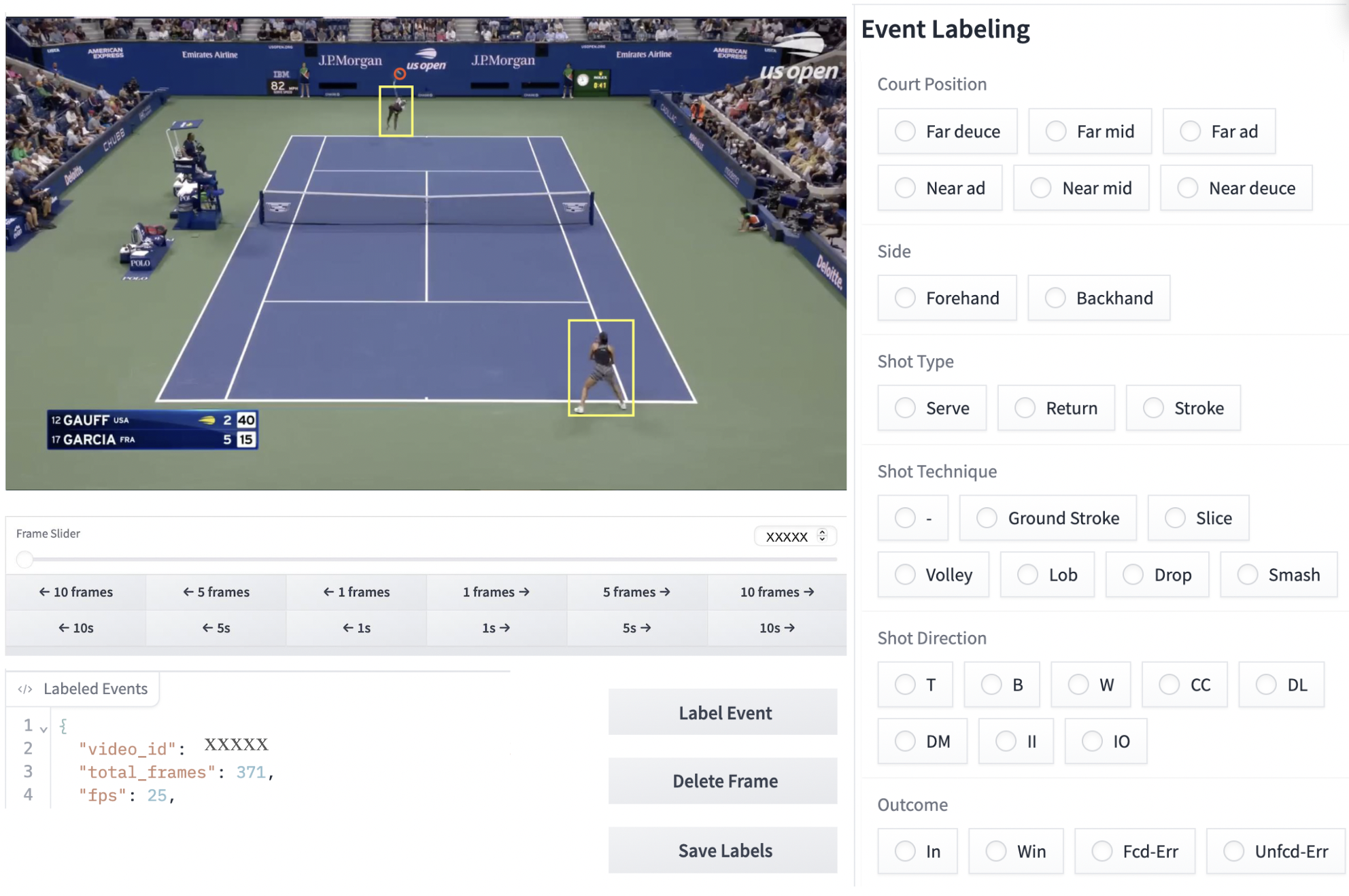}
    \vspace{-10px}
\caption{{\color{black}An interface of the labeling tool. The panel on the right is application-customizable.}}
\vspace{-10px}
\label{fig:interface}
\end{figure}

{\color{black} \paragraph{Annotation pipeline and toolchain.} 
% ADD A PIPELINE!!!
After data collection, we use a three-stage annotation process designed to maximize automation and minimize manual effort. This pipeline is adaptable to various sports broadcast videos and broader domains:}
\vspace{-7px}
{\color{black} 
\begin{description}[leftmargin=*]
    \item \emph{(1) Video segmentation}: The first stage is to segment a full broadcast video into shorter clips using a context-aware scene detector [\citenum{pyscenedetect}] that automatically identifies jump cuts within the video. 
    \vspace{-3px}
    % By detecting areas where the difference between two consecutive frames surpasses a predetermined threshold, we can effectively identify transitions between scenes.
    \item \emph{(2) Clip selection}: The second stage is to select targeted clips (e.g., clips contain tennis rallies) using a Siamese network to compare each clip with a ``base image'' indicative of the scene of interest.
    \vspace{-3px}
    % . Given that the camera typically films a rally from a fixed height and angle, we will first select a ``base image'' that represents the rally scene. After that, we compare the similarity between the ``base image'' and the middle frame of each clip through a Siamese network. If the similarity score surpasses a predetermined threshold, we will keep this clip for further analysis. 
    \item \emph{(3) \fff{} event annotation}: The final stage is to identify the precise event moments (e.g., frames when a player hits the ball) and record the corresponding event types through an annotation tool. 
\end{description}
}

\vspace{-5px}
The first two steps are automated and applicable to a range of sports videos, facilitating the efficient breakdown of lengthy videos into relevant clips.
% The first stage is to segment a full broadcast video into shorter clips. We utilize a context-aware scene detector [\citenum{pyscenedetect}] that identifies jump cuts within the video. By detecting areas where the difference between two consecutive frames surpasses a predetermined threshold, we can effectively identify transitions between scenes.  However, these clips often include extraneous content like audience reactions and player preparations beyond actual tennis rallies.
% Therefore, the second stage is to select clips that contain tennis rallies (i.e., sequences of back-and-forth shots between players within a point). Given that the camera typically films a rally from a fixed height and angle, we will first select a ``base image'' that represents the rally scene. After that, we compare the similarity between the ``base image'' and the middle frame of each clip through a Siamese network. If the similarity score surpasses a predetermined threshold, we will keep this clip for further analysis. 
% Annotators manually review these selected clips to ensure each contains a complete tennis rally, accurately capturing the starting and ending moments and all shots.
{\color{black} For the final phase, we developed an interactive annotation interface, shown in Figure~\ref{fig:interface}. The tool allows users to navigate clips quickly (e.g., 1-second increments) or review them frame by frame, enabling efficient identification of key events (e.g., hitting moments). It supports selecting shot types and identifying court positions through direct clicks on the video, with each click displayed for immediate verification. Object-level detection can assist the process, and a foolproof design minimizes errors from accidental clicks or misjudgments. This tool is adaptable to other sports by incorporating domain-specific knowledge, broadening its applicability.}

% For the final phase, we have developed an interactive interface to support manual annotation, as demonstrated in Figure~\ref{fig:interface}. This tool enables users to navigate through each clip with options for rapid advancement (e.g., in 1-second increments) or more detailed review (e.g., frame by frame), facilitating quicker identification of key event moments (e.g., tennis hitting moments). Additionally, it features functionality for selecting shot types and identifying location data (e.g., court position) by directly clicking on the video. The location of each click is concurrently displayed on the video for immediate verification. Object-level detection can be applied to facilitate annotation. Furthermore, to reduce labeling errors, our labeling tool also incorporates a fool-proof design to avoid unintended manual clicking errors or misjudgment.
% % Although detailed discussions of our script were omitted due to space limitations, we are ready to provide comprehensive details and code examples if required. 
% This tool is adaptable to other sports by integrating domain-specific knowledge and can be applied to broader applications.

% In the final stage, we manually record detailed match data, including scores (set, game, and point) and player specifics such as names and handedness. After that, we iterate through each clip, manually identify the frames when a player hits the ball (the exact moment of the racket contact the ball) and record the corresponding event types for each shot. 

Our annotation team consists of 8 members. We provided them with specialized training and rigorous pre-tests before beginning the official annotation work, along with supporting materials such as slides and demonstrations. Each annotator was assigned an equal portion of the dataset, totaling 1,450 clips (rallies) each. The manual labeling takes roughly 30 hours to finish all 1,450 clips. 
Following the initial annotation phase, we conducted multiple rounds of cross-validation involving random sampling of rallies and quality checks among annotators to ensure the accuracy of the event-based labels. 
% Conflicting samples were resolved using a majority vote criterion.
{\color{black} In cases where conflicting annotations arose, annotators were asked to input the labels they believed to be correct. The final label was determined based on a majority vote among the annotators.}

% % \paragraph{Player bounding boxes.} 
% In a single tennis match, both players occupy a small proportion of the entire scene. 
% % Their subtle movements are crucial for recognizing fine-grained action sequences. 
% Important visual details might be lost for \fff{} event detection after resizing entire images to 224 $\times$ 224. 
% % Therefore, in addition to extracting global features from the entire scene, we also pay attention to local features with rich semantic information, such as areas around the players (e.g., the second row in Figure~\ref{fig:vid2seq}). Such regional information can be used to complement the global semantic feature. 
% Therefore, we also provide the player bounding boxes in each video clip serving as the complement of global features extracted from entire scenes.

% \paragraph{Quality control.} To guarantee the integrity of our dataset with accurate annotations throughout all event types, we adopted comprehensive quality control strategies. These strategies encompassed training annotators with specialized domain knowledge, administering stringent pretests prior to the official annotation process, supplying reference materials such as slides and demonstrations, and enforcing cross-validation procedures among annotators.
\vspace{-10px}

{\color{black}\subsection{\dataset{} dataset statistics and properties} \label{properties}
% We search for tennis single matches from Grand Slams, Olympics, and other major ATP (Association of Tennis Players) and WTA (Women's Tennis Association) tournaments on YouTube and download competition videos in high-resolution. The videos contain different court types (i.e., hard/clay/grass courts), 
% The $\dataset{}$ dataset contains 114 full broadcast tennis single matches with high resolutions and video frame rates ranging from 25 to 30 FPS, involving 75 professional players. Among all the players, 30 are males and 45 are females. Besides, 68 are right handed and 7 are left-handed. From these source videos, we have collected 11,584 video clips, where each clip contains a tennis rally with an average duration of 8.4 second. The total number of shots in each rally ranges from 1 to 34. The dataset contains 42,846 shots. 
\vspace{-5px}
Key statistics for $\dataset{}$ tennis dataset are summarized in Table~\ref{tab:dataset_stats}. Statistics for other \fff{} datasets, including badminton, table tennis, and tennis doubles, are provided in the Appendix~\ref{stats}.} We employ a training, validation, and testing split of 3:1:1, with the training and validation sets drawn from the same video sources, while the test set features clips from distinct videos.  

% The $\dataset{}$ tennis dataset consists of 114 high-resolution broadcast tennis singles matches featuring 75 professional players (30 men and 45 women) with frame rates ranging from 25 to 30 FPS. Among these players, 68 are right-handed, and 7 are left-handed. 
% This dataset encompasses 11,584 video clips, each capturing a tennis rally with an average duration of approximately 8.4 seconds. These clips collectively contain 42,846 tennis shots, with each rally comprising between 1 to 34 shots.

% Additionally, $\dataset{}$ also contains 10 tennis doubles matches, 10 badminton matches, and 5 table tennis matches.
% We employ a training, validation, and testing split of 3:1:1, with the training and validation sets drawn from the same video sources, while the test set features clips from distinct videos.  
% Table~\ref{tab:sports_datasets} presents a more detailed breakdown of this data, highlighting its fast, frequent, and fine-grained nature in comparison to other video datasets. 
% Additional statistical insights can be found in Appendix~\ref{stats}.
\vspace{-5px}

\begin{table}
\small
\centering
\caption{Summary of \dataset{} tennis dataset statistics.}
\vspace{-1.5mm}
\begin{minipage}{.48\linewidth}
\centering
{\color{black}
\begin{tabular}{lc}
\toprule
Category           & Details                 \\ 
\midrule
Matches           & 114 broadcast matches    \\
Players          & 75 (30 men, 45 women)            \\
Handedness       & 68 right-handed, 7 left-handed   \\
Frame Rate (FPS) & 25--30 FPS                  \\
\bottomrule
\end{tabular}
}
% \subcaption{General Statistics}
\label{tab:dataset_stats_1}
\end{minipage}
\hfill
\begin{minipage}{.48\linewidth}
\centering
{\color{black}
\begin{tabular}{lc}
\toprule
Category            & Details                 \\ 
\midrule
Clips              & 11,584 rallies          \\
Avgerage Clip Duration & 8.4 sec                 \\
Total Shots        & 42,846                  \\
Shots Per Rally    & 1 to 34                 \\
\bottomrule
\end{tabular}
}
% \subcaption{Shot Statistics}
\label{tab:dataset_stats}
\end{minipage}
\vspace{-10px}
\end{table}

% \begin{table}
% \small
% \centering
% \caption{Summary of \dataset{} tennis dataset statistics.}
% \vspace{-1.5mm}
% {\color{black}
% \begin{tabular}{lc}
% \toprule
% {Category}        & {Details}                     \\ 
% \midrule
% Matches                 & 114 broadcast matches                \\
% Players                 & 75 (30 men, 45 women)                \\
% Handedness              & 68 right-handed, 7 left-handed       \\
% Frame Rate              & 25--30 FPS                           \\
% Clips                   & 11,584 rallies                       \\
% Average Clip Duration   & 8.4 seconds                          \\
% Total Shots             & 42,846                               \\
% Shots Per Rally         & 1 to 34                              \\ 
% \bottomrule
% \end{tabular}
% }
% \vspace{-10px}
% \label{tab:dataset_stats}
% \end{table}
% \normalsize

\paragraph{Event Timestamp.} Unlike typical TAL and TAS tasks, where an action spans several frames or seconds, the duration of actions in racket sports is often ambiguous. Thus, stroke actions are defined as instantaneous events, recording only the moment of ball-racket contact [\citenum{voeikov2020ttnet}] as shown in Figure~\ref{fig:vid2seq}.
\vspace{-8px}

\paragraph{Multi-level granularity.} 
{\color{black} Depending on the requirements of the analytics task, \dataset{} can focus on a subset of sub-classes, enabling flexible granularity. We define a parameter \( G \in \mathcal{P}(\{sc_1, \ldots, sc_8\}) \), where \( \mathcal{P}(\{sc_1, \ldots, sc_8\}) \) is the power set of \( \{sc_1, \ldots, sc_8\} \), to select sub-classes and form different levels of granularity. We define 3 granularity levels using \dataset{} tennis as an example.
At the coarse level, \( G_{\text{low}} = \{sc_1, sc_3, sc_4, sc_8\} \) includes 4 sub-classes, 11 elements, and 38 event types. This level captures essential but broad information.
At a finer level, \( G_{\text{mid}} = \{sc_1, \ldots, sc_6\} \) consists of 6 sub-classes, 24 elements, and 365 event types. This granularity provides more detailed event representations.
At the most detailed level, \( G_{\text{high}} = \{sc_1, \ldots, sc_8\} \) encompasses all 8 sub-classes, 29 elements, and 1,108 event types. This level is ideal for precise and comprehensive event analysis.
% This fine-grained level is ideal for tasks demanding precise and comprehensive event analysis.
% \begin{description}[leftmargin=*]
%     \item At the coarse level, \( G_{\text{low}} = \{sc_1, sc_3, sc_4, sc_8\} \) includes 4 sub-classes, 11 elements, and 38 event types. This level captures essential but broad information.
%     \item At a finer level, \( G_{\text{mid}} = \{sc_1, \ldots, sc_6\} \) consists of 6 sub-classes, 24 elements, and 365 event types. This granularity provides more detailed event representations.
%     \item At the most detailed level, \( G_{\text{high}} = \{sc_1, \ldots, sc_8\} \) encompasses all 8 sub-classes, 29 elements, and 1,108 event types. This fine-grained level is ideal for tasks demanding precise and comprehensive event analysis, such as detailed tennis strategy analytics.
% \end{description}
This multi-level granularity enhances \dataset{}'s flexibility for diverse real-world tasks. 
\vspace{-8px}

\subsection{Ethical Considerations}
\vspace{-5px}
\dataset{} is constructed from publicly available sports broadcasts, ensuring compliance with ethical and legal standards. We do not redistribute video content, providing only YouTube links to maintain adherence to copyright policies. The dataset focuses on professional players in public tournaments, avoiding private or off-court data and ensuring it is used strictly for academic research. While anonymization is not applied, as these players are public figures, we emphasize that the dataset should not be used for non-research purposes. 
% We periodically update the dataset, ensuring ongoing availability and addressing content removals responsibly. 
A more detailed discussion on privacy, consent, and bias mitigation is provided in Appendix~\ref{ethical}.
}
\vspace{-5px}

\section{Our Proposed Approach: $\fest{}$} \label{our_method}
\begin{figure}%[t]
\centering
    \includegraphics[width=0.85\linewidth]{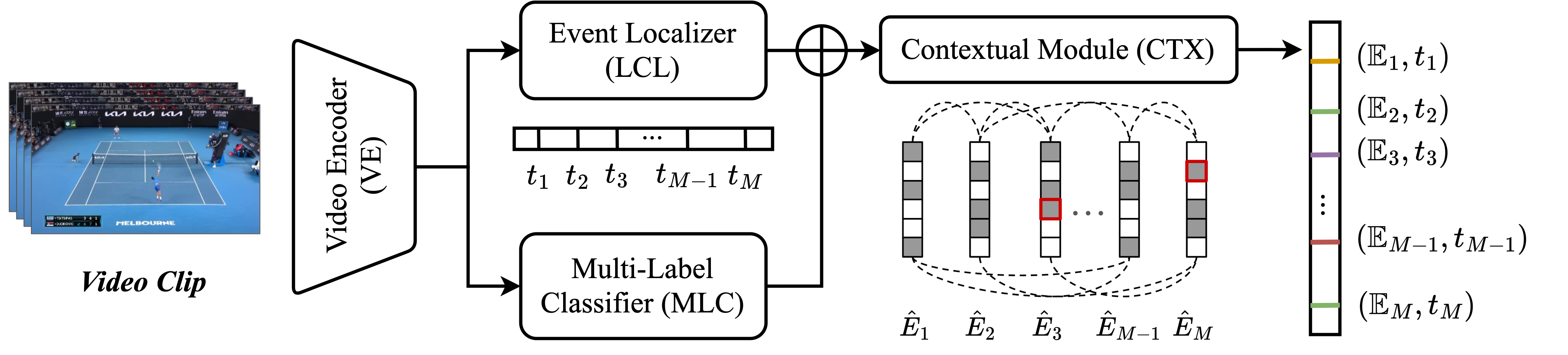}
    \vspace{-5px}
\caption{{\color{black} Overview of $\fest{}$. RGB images are processed by VE to capture frame-wise spatial-temporal features, which are passed to LCL to identify event timestamps and MLC to predict labels. Outputs from LCL and MLC are combined (`plus' symbol) to form an event representation sequence and refined by CTX module. `Red squares' represent errors from purely visual predictions.}}
\vspace{-15px}
\label{fig:arch}
\end{figure}

\vspace{-5px}
Acknowledging the challenges and limitations of existing approaches, we propose a simple yet effective method named \textbf{F}ast \textbf{F}requent \textbf{F}ine-grained \textbf{E}vent \textbf{D}etection network (\fest{}), illustrated in Figure~\ref{fig:arch}. It is designed for \fff{} event detection and can serve as a baseline for further development. 
\vspace{-5px}

\paragraph{Problem formulation.} Let $X \in \mathbb{R}^{H \times W \times 3 \times N}$ denote the input, consisting of $N$ RGB frames of size $H \times W$. The output is a sequence of $M$ event-timestamp pairs $((E_1,t_1), \ldots, (E_M,t_M))$, where $E_i$ is the event type with $C$ classes and $t_i$ is the corresponding timestamp for $i \in \{1, \ldots, M\}$. Additionally, each event $E_i$ can also be expressed as a vector $[e_{i,1}, \ldots, e_{i,K}]$, with each element $e_{i,j} \in \{0, 1\}$ indicating the presence or absence of the {\color{black} $j^{th}$ element} in event $E_i$, {\color{black} where $j$ is an integer $j \in \{1, \ldots, K\}$}. The parameter $K$, which defines the \emph{number of elements} in each event vector.
\vspace{-5px}

% , varies with the dataset granularity, being set to $K=11$ for $G_{low}$, $K=24$ for $G_{mid}$, and $K=29$ for $G_{high}$. 

% \paragraph{Overall Architecture.} \fest{} comprises four main components, as depicted in Figure~\ref{fig:arch}: (1) a video encoder that extracts spatial-temporal features, (2) an event detector designed to accurately pinpoint the timestamps of critical events, (3) a multi-label event classifier that leverages visual features at the identified timestamps, and (4) a causality module that models contextual knowledge within detected event sequences.

% \subsection{Video feature encoding}
\paragraph{Video Encoder (VE).} The first stage of both baselines and our model will extract spatial-temporal frame-wise features. The video encoder (VE) consists of a visual backbone, followed by a bidirectional GRU to capture long-term visual dependencies: $\textbf{F}_{emb} = \text{VE}(X)$, with $\textbf{F}_{emb} \in R^{N \times d'}$.

\vspace{-10px}
% \subsection{Fine-grained event detection}
\paragraph{Event Localizer (LCL).} Utilizing the frame-wise features \(\textbf{F}_{emb}\), the event localizer (LCL) employs a fully connected network with a Sigmoid activation function to perform dense binary classification, aiming to accurately identify specific event instances. For an \(N\)-frame clip, the output is represented as \((\hat{p}_1, \ldots, \hat{p}_N)\), where each \(\hat{p}_i\) denotes the probability that an event occurs at the corresponding timestamp: \((\hat{p}_1, \ldots, \hat{p}_N) = \textit{Sigmoid}(\text{LCL}(\textbf{F}_{emb}))\). Ground truth labels \((p_1, \ldots, p_N)\) with \(p_i \in \{0, 1\}\) are used to compute the discrepancy between the predicted probabilities and the actual values using binary cross-entropy loss as:
$
% \begin{equation}
    L_{LCL} = \frac{1}{N}\sum^{N}_{i=1}p_i \cdot log(\hat{p}_i) + (1 - p_i) \cdot log(1 - \hat{p}_i).
% \end{equation}
$
\vspace{-5px}

% \paragraph{Multi-label event classifier (MLC).} After we have detected the events, we further classify the shots into fine-grained event types using a \emph{multi-label} classification module (MLC). The MLC is a fully connected network that takes the spotted event features as input and predicts the event types: $\hat{E}_i = \textit{Sigmoid}(\text{MLC}(f_i)) = [\hat{e}_{i,1}, \ldots, \hat{e}_{i,K}],$
% where $K$ is the number of elements, $f_i \in \textbf{F}_{emb}$ is the feature for the event instance at $i^{th}$ frame, $\hat{E}_i$ is the predicted event type, and $\hat{e}_{i,j} \in [0, 1]$ is the probability of $\hat{E}_i$ containing the $j^{th}$ element. For a video clip with $M$ events and ground truths $(E_1, \ldots, E_M)$, where $E_i$ is represented as a vector of $K$ elements $[e_{i,1}, \ldots, e_{i,K}]$.
\paragraph{Multi-label Event Classifier (MLC).} Upon detecting events, we proceed to categorize them into specific types using a multi-label classification module (MLC). This module, a fully connected network, takes the identified event features \(f_i\) from \(\textbf{F}_{emb}\) as inputs to predict the event types: \(\hat{E}_i = \textit{Sigmoid}(\text{MLC}(f_i)) = [\hat{e}_{i,1}, \ldots, \hat{e}_{i,K}]\), where \(K\) denotes the number of elements, \(f_i\) represents the features for the event at the \(i^{th}\) frame, \(\hat{E}_i\) is the predicted event type, and \(\hat{e}_{i,j} \in [0, 1]\) is the probability of \(\hat{E}_i\) containing the \(j^{th}\) element. For a video clip with \(M\) events, the ground truths are given as \((E_1, \ldots, E_M)\) with each \(E_i\) represented as a vector of \(K\) elements \([e_{i,1}, \ldots, e_{i,K}]\). The loss can be represented by
$
% \begin{equation}
    L_{MLC} = \frac{1}{M} \sum^{M}_{i=1} (\frac{1}{K} \sum^{K}_{j=1}e_{i,j} \cdot log(\hat{e}_{i,j}) + (1 - e_{i,j}) \cdot log(1 - \hat{e}_{i,j})).
% \end{equation}
$
\vspace{-10px}

% It is often challenging for video encoders to capture insightful visual features from fast-paced videos due to motion blur. Moreover, the objects of interest (e.g., players) may occupy a small fraction of the entire scene, which will lead to the loss of important visual details for fine-grained action classification when resizing images to 224 $\times$ 224. Consequently, naively selecting the best-predicted event types might yield invalid event sequences. Therefore, we introduce a contextual module (CTX), which concurrently learns contextual knowledge from event sequences during end-to-end training: $(\mathbb{E}_1, \ldots, \mathbb{E}_M) = \text{CTX}(\hat{E}_1, \ldots, \hat{E}_M),$ where CTX is a bidirectional GRU that processes predicted event sequence $\hat{E}$ and outputs a refined one $\mathbb{E}_i = [\mathbbm{e}_1, \ldots, \mathbbm{e}_k]$ by considering both visual-based predictions and contextual correlations across events. 
\paragraph{Contextual module (CTX)} \label{sec:cal}
Video encoders often struggle to extract insightful visual features from fast-paced videos due to motion blur, and objects of interest, such as players, may only occupy a small portion of the frame. This can result in the loss of crucial visual details for fine-grained action classification, particularly when resizing images to 224 \(\times\) 224. Selecting the best-predicted event types naively might, therefore, produce invalid event sequences. To address this, we introduce a contextual module (CTX), designed to concurrently learn contextual knowledge from event sequences during end-to-end training: \((\mathbb{E}_1, \ldots, \mathbb{E}_M) = \text{CTX}(\hat{E}_1, \ldots, \hat{E}_M)\). CTX employs a bidirectional GRU to process the predicted event sequence \(\hat{E}\) and outputs a refined sequence \(\mathbb{E}_i = [\mathbbm{e}_1, \ldots, \mathbbm{e}_k]\), integrating both visual-based predictions and contextual correlations across events.
The loss is calculated for each refined event:
$
% \begin{equation}
    L_{CTX} = \frac{1}{M} \sum^{M}_{i=1} (\frac{1}{K} \sum^{K}_{j=1}e_{i,j} \cdot log(\mathbbm{e}_{i,j}) + (1 - e_{i,j}) \cdot log(1 - \mathbbm{e}_{i,j})).
% \end{equation}
$
\vspace{-10px}

\section{Experiments} \label{sec:experiment}
\vspace{-5px}
% This section describes our experimental methods for benchmarking with existing works and ablation studies. 
In this section, we benchmark existing temporal action understanding methods, including TAL, TAS, and TASpot, on the \dataset{} dataset and conduct a series of ablation studies.
\vspace{-5px}

% \paragraph{Problem formulation.} Let $X \in \mathbb{R}^{H \times W \times 3 \times N}$ denote the input, consisting of $N$ RGB frames of size $H \times W$. The output is a sequence of $M$ event-timestamp pairs $((E_1,t_1), \ldots, (E_M,t_M))$, where $E_i$ is the event type and $t_i$ is the corresponding timestamp for $i \in \{1, \ldots, M\}$. Additionally, each event $E_i$ can be expressed as a vector $[e_{i,1}, \ldots, e_{i,K}]$, with each element $e_{i,j} \in \{0, 1\}$ indicating the presence or absence of the $j^{th}$ element in event $E_i$, for $j \in \{1, \ldots, K\}$. The parameter $K$, which defines the \emph{number of elements} in each event vector, varies with the dataset granularity, being set to $K=11$ for $G_{low}$, $K=24$ for $G_{mid}$, and $K=30$ for $G_{high}$. 
% The parameter $K=30$ is specified in the dataset $\dataset{}$, as detailed in Section~\ref{sec:dataset}. 

% \paragraph{Evaluation metrics.}
% We use two types of evaluation metrics: (1) sequence-wise edit score [\citenum{lea2017temporal}], and (2) mean F1 score within a tolerance of $\pm1$ frame. 
% We use mean F1 score within a tolerance of $\pm1$ frame and sequence-wise edit score [\citenum{lea2017temporal}] as evaluation metrics.
% \vspace{-5px}
{\color{black} \paragraph{Evaluation metrics.} The evaluation metrics used in our work are carefully chosen to comprehensively assess both the temporal precision and classification accuracy of detected events, which are critical for \fff{} event detection. These metrics align with evaluation standards in similar tasks.
\textbf{Edit Score} [\citenum{lea2017temporal}] measures the similarity between predicted and ground truth event sequences using Levenshtein distance, capturing errors in event sequence structure, such as missing, additional, or misordered events. This metric is particularly valuable for evaluating models where the temporal order and completeness of event sequences are essential [\citenum{he2024vistec}].
\textbf{Mean F1 Score with Temporal Tolerance} evaluates both classification and temporal localization accuracy [\citenum{hong2022spotting,he2024vistec}]. By considering a prediction correct only when its timestamp aligns within a strict temporal tolerance (e.g., $\pm1$ frame) and its class correctly identifies, this metric ensures that models are assessed on their ability to achieve precise temporal spotting alongside accurate classification. Given the long-tail distribution of event types in the dataset, where some events 
are extremely rare, we report two variants of the mean F1 score to ensure a balanced evaluation: $F1_{evt}$, the average F1 score across all event types, and $F1_{elm}$, the average F1 score across all elements, which typically presents a more balanced distribution.
% The two variants, $F1_{evt}$ and $F1_{elm}$, further enrich the evaluation by focusing on event types and elements, respectively, addressing class imbalance and providing a balanced assessment across diverse event distributions.
% Together, these metrics provide a robust framework to evaluate the temporal, sequential, and classification aspects of \fff{} event detection.
}
\vspace{-5px}

\paragraph{Baselines.} Existing temporal action understanding frameworks typically incorporate two key components: a \emph{video encoder} for visual feature extraction and a \emph{head module} for specific tasks such as detection or segmentation. Applying these models directly to our study presents challenges, as they generally utilize a two-stage training process—employing a static, pre-trained video encoder for feature extraction and training only the head module. This approach often fails to capture fine-grained, domain-specific events due to its reliance on temporally coarse, non-overlapping, or downsampled video segments.
To address these limitations, we have adapted these {\color{black}temporal action understanding methods} to develop new baselines better suited for detecting \fff{} events. Given the rapid pace and short duration of tennis shots, it is crucial to utilize frame-wise feature extraction [\citenum{chen2022frame}] {\color{black}(discussed in Section~\ref{sec:ablation})}. Besides, end-to-end training with video encoder fine-tuning is required to capture the subtle event differences. Moreover, the classification of some sub-classes (e.g., shot direction, outcome) demands long-term temporal reasoning to integrate information from subsequent frames.

Consequently, we focus on established feature extractors: {TSN} [\citenum{wang2018temporal}], {SlowFast} [\citenum{feichtenhofer2019slowfast}], I3D [\citenum{carreira2017quo}], VTN [\citenum{neimark2021video}], and {TSM} [\citenum{lin2019tsm}], which enable 
% are known for their efficiency in 
frame-wise feature extraction and end-to-end training. We pair each encoder with five representative head module architectures from existing methods: MS-TCN [\citenum{farha2019ms}] and ASFormer [\citenum{yi2021asformer}] from TAS, G-TAD [\citenum{xu2020g}] and ActionFormer [\citenum{zhang2022actionformer}] from TAL and E2E-Spot [\citenum{hong2022spotting}] from TASpot, to establish a set of new baseline models for our study. To identify hitting moments and their respective event types, frame-wise dense \emph{multi-class} classification is applied to identify each frame as either background or one of the event types.
\vspace{-5px}

\vspace{-5px}
\paragraph{Implementation details.} We implement and train models on \dataset{} in an end-to-end manner. The video encoder takes video clip $X$ down-scaled and cropped to 224 $\times$ 224 to extract frame-wise visual features. 
Subsequently, each head module processes per-frame features to identify a sequence of \fff{} events and their timestamps. For more implementation details, please refer to Appendix~\ref{implement}.
\vspace{-5px}
% \paragraph{Implementation details.} We implement and train models on \dataset{} in an end-to-end manner. The video encoder takes video clip $X$ down-scaled and cropped to 224 $\times$ 224 to extract frame-wise visual features $\textbf{F}_{vid} \in \mathbb{R}^{N \times d}$. 
% Subsequently, each head module processes $\textbf{F}_{vid}$ to produce encoded features $\textbf{F}_{enc} \in \mathbb{R}^{N \times d'}$. To identify hitting moments and their respective event types, frame-wise dense classification is applied on $\textbf{F}_{enc}$, resulting in predictions for each frame $P=(p_1, \ldots, p_N)$, where $p_i$ is classified as either background or one of the event types. The cross-entropy loss is computed as
% $
% % \begin{equation}
%     L = \frac{1}{N} \sum^{N}_{i=1} \text{CrossEntropy}(p_i, y_i),
% % \end{equation}
% $
% where $y_i$ represents the ground truth for frame $i$. For more details and our \fest{}, please refer to Appendix~\ref{implement}.

\begin{table}
\caption{
Experimental results on \dataset{} (tennis) with 3 levels of granularity. Full table in Appendix~\ref{results_sup}.
% The best and second-best results are shown in bold and underlined, respectively. 
} 
\vspace{-1.5mm}
\centering
\small
\resizebox{\textwidth}{!}{
\begin{tabular}{@{}llccccccccccc@{}}
\toprule
 & & \multicolumn{3}{c}{$\dataset{}\,(G_{high})$} & \multicolumn{3}{c}{$\dataset{}\,(G_{mid})$} & \multicolumn{3}{c}{$\dataset{}\,(G_{low})$} \\
\cmidrule(r){3-5} \cmidrule(lr){6-8} \cmidrule(l){9-11}
Video encoder & Head arch. & F$1_{evt}$ & F$1_{elm}$ & Edit & F$1_{evt}$ & F$1_{elm}$ & Edit & F$1_{evt}$ & F$1_{elm}$ & Edit \\
\midrule
TSN [\citenum{wang2018temporal}]
&MS-TCN [\citenum{farha2019ms}]
&15.9 & 59.8 & 53.5 &23.2 & 60.9 & 65.8 &45.7 & 70.4 & 72.8 \\
% &ASformer  [\citenum{yi2021asformer}]
% &11.9 & 54.3 & 49.8 &17.3 & 56.1 & 62.5 &40.3 & 67.3 & 70.3 \\
% &G-TAD [\citenum{xu2020g}]
% &6.0 & 47.5 & 24.7 &14.1 & 52.1 & 48.6 &19.9 & 57.4 & 44.7 \\
&ActionFormer [\citenum{zhang2022actionformer}]
&18.4 & 60.6 & 55.2 &24.8 & 61.9 & 67.3 &48.7 & 70.6 & 72.2 \\
&E2E-Spot [\citenum{hong2022spotting}]
&24.7 & 65.3 & 60.1 &31.5 & 66.2 & 71.0 &53.5 & 73.6 & 75.0 \\
\midrule
SlowFast [\citenum{feichtenhofer2019slowfast}]
% &MS-TCN [\citenum{farha2019ms}]
% &17.2 & 63.1 & 56.2 &24.3 & 65.5 & 70.3 &47.4 & 73.1 & 73.5 \\
% &ASformer  [\citenum{yi2021asformer}]
% &14.1 & 60.8 & 55.3 &20.3 & 62.8 & 69.4 &44.8 & 72.9 & 71.9 \\
&G-TAD [\citenum{xu2020g}]
&23.0 & 66.1 & 64.0 &29.6 & 66.5 & 74.2 &53.3 & 76.0 & 77.9 \\
&ActionFormer [\citenum{zhang2022actionformer}]
&28.7 & {70.0} & {67.6} &35.5 & {70.9} & {76.4} &59.3 & {77.1} & {81.5} \\
&E2E-Spot [\citenum{hong2022spotting}]
&25.9 & 69.4 & 65.7 &33.8 & 70.4 & 75.4 &55.5 & 76.5 & 79.5 \\
\midrule
{\color{black}  I3D [\citenum{carreira2017quo}]} & {\color{black}E2E-Spot [\citenum{hong2022spotting}]} &{\color{black}22.7} & {\color{black}59.7} & {\color{black}68.7} & {\color{black}27.1} & {\color{black}60.7} & {\color{black}74.2} & {\color{black}51.9} & {\color{black}67.7} & {\color{black}78.3} \\
{\color{black}  VTN [\citenum{neimark2021video}]} & {\color{black}E2E-Spot [\citenum{hong2022spotting}]} &{\color{black}14.8} & {\color{black}58.3} & {\color{black}56.7} & {\color{black}20.0} & {\color{black}59.4} & {\color{black}68.2} & {\color{black}39.7} & {\color{black}63.1} & {\color{black}73.1}\\
\midrule
TSM [\citenum{lin2019tsm}]
&MS-TCN [\citenum{farha2019ms}]
& 21.7 & 67.3 & 58.6 & 30.4 & 69.5 & 73.0 &50.2 & 74.0 & 75.3 \\
&ASformer  [\citenum{yi2021asformer}]
& 17.6 & 61.9 & 57.5 & 25.5 & 64.0 & 74.2 &46.0 & 72.9 & 74.0\\
&G-TAD [\citenum{xu2020g}]
& 16.9 & 62.5 & 55.2 & 29.8 & 66.9 & 74.8 & 39.8 & 70.1 & 67.2 \\
&ActionFormer [\citenum{zhang2022actionformer}]
& 22.4 & 65.7 & 60.3 & 31.0 & 68.2 & 74.7 & 52.4 & 73.8 & 74.9 \\
&E2E-Spot [\citenum{hong2022spotting}]
& {31.4} & {71.4} & {68.7} & 39.5 & {72.3} & {77.9} & 60.6 & {78.4} & {82.1} \\
\midrule
TSM[\citenum{lin2019tsm}] &{$\fest{}$} &\textbf{40.3} & \textbf{75.2} & \textbf{74.0} & \textbf{48.0} & \textbf{76.5} & \textbf{82.4} &\textbf{68.4} & \textbf{80.0} & \textbf{87.2} \\
\bottomrule
\end{tabular}
}
\vspace{-20px}
\label{tab:results}
\end{table}
\normalsize

\vspace{-5px}
\subsection{Results and analysis} \label{result&analysis}

\vspace{-5px}
{\color{black}The evaluation results presented in Table~\ref{tab:results} provide several critical insights into the performance of various methods across different levels of granularity ($G_{low}$, $G_{mid}$, and $G_{high}$). A general trend emerges where performance decreases as granularity increases, underscoring the growing challenges associated with finer granularity. While certain methods demonstrate some robustness, the overall efficacy across all approaches remains suboptimal, particularly at higher levels of granularity, indicating the challenge of precise \fff{} event detection task.

% Simple 2D CNNs such as TSN, which process frames independently, are inadequate for \fff{} event detection due to their inability to capture critical spatial-temporal correlations between frames, which are essential for distinguishing visually similar events. Without modeling temporal dynamics, these approaches struggle to differentiate events that may appear identical when viewed frame-by-frame, leading to significantly lower performance, particularly at higher granularity levels.
Simple 2D CNNs (e.g., TSN), which process frames independently, are inadequate for \fff{} event detection due to their inability to capture critical spatial-temporal correlations between frames. Lacking temporal modeling, they struggle to distinguish visually similar events, resulting in poor performance, especially at higher granularity levels.
Advanced video encoders such as I3D [\citenum{carreira2017quo}], SlowFast [\citenum{feichtenhofer2019slowfast}], and transformer-based VTN [\citenum{neimark2021video}], which excel in other video understanding tasks, face significant challenges with \dataset{}. These models process video data using techniques like non-overlapping snippets or frame downsampling, resulting in coarse temporal features. While effective for long-duration actions, such approaches struggle to detect the rapid, short-duration events in \fff{}, which rely on precise temporal cues spanning only 1–2 frames. This suggests that increasing video encoder complexity does not necessarily improve performance for fast-action detection in \dataset{}. Notably, simpler models like TSM, paired with advanced 2D CNNs such as RegNet-Y [\citenum{radosavovic2020designing}], outperform these complex encoders. This highlights the importance of capturing subtle visual differences over short temporal spans, demonstrating that the ability to extract fine-grained temporal cues is more impactful than model complexity.

% Interestingly, TSM combined with E2E-Spot outperforms the more complex SlowFast model, indicating that the complexity of video encoder might not scale well to the fast action detection performance of \dataset{}. Instead, it is more important to capture subtle visual differences over short temporal durations, which are crucial for \fff{} event detection. This result suggests that the capability of capturing subtle temporal cues and representation is more impactful than the model complexity.

Head modules such as transformer-based ActionFormer, and GRU-based E2E-Spot, generally outperform other methods. This advantage highlights their effectiveness in capturing long-term temporal dependencies through end-to-end training. 
% By integrating spatial and temporal features seamlessly, these models demonstrate enhanced ability to recognize complex event patterns. 
Notably, E2E-Spot consistently outperforms ActionFormer across most settings, suggesting that GRU-based architectures may offer an advantageous trade-off between efficiency and representational power for certain types of temporal correlations.

Our proposed \fest{} model, leveraging the TSM video encoder, achieves the best performance among all granularity levels. This is attributable to two key design choices: the multi-label classifier and the contextual module. Detailed discussions of these design elements are presented in the next section.}
\vspace{-5px}

% Beyond the statistical results in Table~\ref{tab:results}, analysis of predicted event sequences reveals that current baselines may produce invalid sequences due to logical errors or uncommon practices. For instance, a right-handed player cannot logically direct a forehand shot from the deuce court as ``II'' or ``IO''. Similarly, an event ending in a winner or error should logically conclude the sequence. Additionally, it is uncommon for a player to hit with backhand when the ball is played to their forehand side. Further examples are detailed in Appendix~\ref{error_sequences}. These observations indicate that existing baselines fail to effectively capture event-wise causal correlations.

\begin{table}
\caption{Ablation and analysis experiments. The default model takes stride size 2 and clip length 96.}
\vspace{-2mm}
\centering
\small
\resizebox{0.95\textwidth}{!}{
\begin{tabular}{@{}lcccccccccccc@{}}
\toprule
& \multicolumn{3}{c}{$\dataset{}\,(G_{high})$} & \multicolumn{3}{c}{$\dataset{}\,(G_{mid})$} & \multicolumn{3}{c}{$\dataset{}\,(G_{low})$} \\
\cmidrule(r){2-4} \cmidrule(lr){5-7} \cmidrule(l){8-10} 
{Experiment}  & F$1_{evt}$ & F$1_{elm}$ & Edit & F$1_{evt}$ & F$1_{elm}$ & Edit & F$1_{evt}$ & F$1_{elm}$ & Edit \\
\midrule
% {\textit{Default (stride 2, clip len 96)}}  &  &  &  &  &  &  &  &  \vspace{1mm}\\
{TSM + E2E-Spot} & {31.4} & {71.4} & {68.7} & 39.5 & {72.3} & {77.9} & 60.6 & {78.4} & {82.1} \\
\midrule
{\color{black}\textit{(a) Feature extractor}}  &  &  &  &  &  &  &  &  \vspace{1mm}\\
{\color{black}  I3D [\citenum{carreira2017quo}] (clip-wise)} &{\color{black}22.7} & {\color{black}59.7} & {\color{black}68.7} & {\color{black}27.1} & {\color{black}60.7} & {\color{black}74.2} & {\color{black}51.9} & {\color{black}67.7} & {\color{black}78.3} \\
{\color{black}  VTN [\citenum{neimark2021video}] (video transformer)} &{\color{black}14.8} & {\color{black}58.3} & {\color{black}56.7} & {\color{black}20.0} & {\color{black}59.4} & {\color{black}68.2} & {\color{black}39.7} & {\color{black}63.1} & {\color{black}73.1}\\
% {\color{black}\textit{(a.2) Skeleton-based}}  &  &  &  &  &  &  &  &  \vspace{1mm}\\
{\color{black}  ST-GCN++ [\citenum{duan2022pyskl}] (skeleton-based)} &{\color{black}25.4} & {\color{black}62.1} & {\color{black}56.1} & {\color{black}32.4} & {\color{black}63.9} & {\color{black}63.5} & {\color{black}55.1} & {\color{black}69.4} & {\color{black}73.2} \\
{\color{black}  PoseConv3D [\citenum{duan2022revisiting}] ( (skeleton-based))} &{\color{black}20.1} & {\color{black}54.5} & {\color{black}53.2} & {\color{black}26.0} & {\color{black}55.4} & {\color{black}61.9} & {\color{black}48.8} & {\color{black}63.0} & {\color{black}69.7} \\
\midrule
% {\textit{(b) Stride size}}  &  &  &  &  &  &  &  &  \vspace{1mm}\\
\textit{(b) Stride size} = 4 &25.9 & 69.2 & 62.7 &33.4 & 69.9 & 73.0 &60.0 & 77.9 & 78.8 \\
\textit{\,\,\,\,\,\,\, Stride size} = 8 &14.0 & 56.7 & 44.3 &18.5 & 57.4 & 54.8 &40.4 & 67.0 & 59.2 \\
\midrule
\textit{(c) without GRU} &27.6 & 69.0 & 60.6 &38.0 & 71.3 & 75.3 &54.7 & 74.1 & 73.4 \\
\midrule
% {\textit{(d) Clip length}}  &  &  &  &  &  &  &  &  \vspace{1mm}\\
% Length = 8 &8.9 & 52.7 & 31.2 &19.7 & 58.0 & 59.7 &23.8 & 60.6 & 47.2 \\
\textit{(d) Clip Length} = 32 &26.3 & 67.4 & 54.5 &35.5 & 69.4 & 71.8 &53.2 & 75.1 & 68.9 \\
\textit{\,\,\,\,\,\,\, Clip Length} = 64 &30.7 & 71.2 & 67.4 &38.6 & 72.4 & 77.5 &58.4 & 77.9 & 81.1 \\
\textit{\,\,\,\,\,\,\, Clip Length} = 192 &29.3 & 70.3 & 65.7 &37.3 & 71.4 & 77.0 &58.8 & 77.1 & 80.4 \\
% \midrule
% {\color{black} \textit{() Resolution (default 224$\times$224)}}  &  &  &  &  &  &  &  &  \vspace{1mm}\\
% 336$\times$336 &43.2 &77.1 &74.8 & &  &  & &  & \\
% 448$\times$448 &44.4 & 78.1 & 74.5 & & & & &  &  \\
% Length = 64 &30.7 & 71.2 & 67.4 &38.6 & 72.4 & 77.5 &58.4 & 77.9 & 81.1 \\
% Length = 192 &29.3 & 70.3 & 65.7 &37.3 & 71.4 & 77.0 &58.8 & 77.1 & 80.4 \\
\midrule
{\textit{(e) Multi-label}} &37.9 & 74.3 & 71.7 & 45.9 & 75.6 & 80.1 &66.6 & 80.1 & 85.1 \\
\midrule
{\color{black} \textit{(f) Multi-label + CTX (Transformer)}} & {\color{black}39.0} & {\color{black}74.3}	& {\color{black}72.8} & {\color{black}50.5} & {\color{black}75.5} & {\color{black}81.8} & {\color{black}63.4} & {\color{black}79.6} & {\color{black}86.8} \\
{\textit{\,\,\,\,\,\,  Multi-label + CTX (BiGRU)}} &\textbf{40.3} & \textbf{75.2} & \textbf{74.0} & \textbf{48.0} & \textbf{76.5} & \textbf{82.4} &\textbf{68.4} & \textbf{80.0} & \textbf{87.2} \\
\bottomrule
\end{tabular}
}
\vspace{-15px}
\label{tab:ablation}
\end{table}
\normalsize

\subsection{Ablation Study} \label{sec:ablation}
\vspace{-5px}
We selected the highest-performing baseline model (TSM + E2E-Spot) as our default configuration for the subsequent ablation studies. {\color{black} More ablation studies can be found in Appendix~\ref{ablation_sup}.}
\vspace{-5px}
\vspace{-5px}
{\color{black}\paragraph{Feature extractor.} An effective feature extractor is crucial for accurate \fff{} event detection. Below, we summarize some key findings (details in Appendix~\ref{ablation_sup}).
% An effective feature extractor is vital for accurate \fff{} event detection. Below, we evaluate various feature extraction methods. Implementation details are detailed in Appendix~\ref{ablation_sup}.
\emph{(1) Frame-wise feature extraction outperforms clip-wise methods}, which divide inputs into non-overlapping segments. Experiments show clip-wise methods produce temporally coarse features and hinder precise event detection.
% Frame-wise feature extraction outperforms clip-wise methods, which divide inputs into non-overlapping segments. Using I3D [\citenum{carreira2017quo}] as a clip-wise extractor on 6-frame segments produces temporally coarse features, resulting in poor performance and emphasizing the importance of frame-level granularity for precise event detection.
\emph{(2) Transformer-based video encoders} such as VTN [\citenum{neimark2021video}] struggle on \dataset{} due to high computational costs and limited ability to effectively capture short-term temporal correlations.
% \emph{(2) Transformer-based encoders:} While transformer-based video encoders like VTN [\citenum{neimark2021video}] offer strong performance, their computational cost limits sequence length and batch size. On \dataset{}, VTN underperforms due to its inability to effectively capture short-term temporal correlations critical for dense sampling.
\emph{(3) In addition to RGB inputs, we also experimented with skeleton-based pose estimation methods}, including ST-GCN++ [\citenum{duan2022pyskl}] and PoseConv3D [\citenum{duan2022revisiting}] with human key points as input. 
% ST-GCN++ outperformed PoseConv3D, but RGB-based features from TSM surpassed both across all granularities. 
While they excel in efficiency and interpretability, they lack critical details like shot direction, limiting performance on \dataset{}.}
\vspace{-5px}

\vspace{-5px}
\paragraph{Sparse sampling.}
Increasing the stride size allows for a broader temporal coverage within a fixed sequence length. This sparse sampling technique is prevalent in many video understanding tasks [\citenum{liu2022empirical,lin2021learning}], offering high efficiency and reasonable accuracy. However, this approach proves inadequate for our task, where events are characterized by their rapid occurrence, frequency, and fine granularity. As illustrated in Table~\ref{tab:ablation}(b), increasing the stride size to 4 and 8 leads to a marked decline in performance, underscoring the importance of dense sampling for detecting \fff{}.
\vspace{-5px}

\vspace{-5px}
\paragraph{Long-term temporal reasoning.}
The default model employs a spatio-temporal video encoder (TSM), complemented by a bidirectional Gated Recurrent Unit [\citenum{dey2017gate}] (GRU) head for enhanced long-term temporal integration. To assess the necessity of long-term temporal reasoning, we replaced the GRU module with a fully connected layer. The results, presented in Table~\ref{tab:ablation}(c), indicate a significant performance decline relative to the original configuration. This finding highlights the essential role of long-term temporal reasoning in analyzing sub-classes such as shot direction, outcomes, and player movements that require information from subsequent frames.
\vspace{-5px}

\vspace{-5px}
\paragraph{Clip length.}
The sensitivity of sequence models to varying input clip lengths, which encapsulate different temporal contexts, is notable. In \dataset{}, the incidence of \fff{} events correlates directly with clip length. Table~\ref{tab:ablation}(d) shows that shorter clips result in fewer events per sequence, hindering the model's ability to leverage long-term dependencies among consecutive events effectively. Conversely, while longer clip lengths yield improved results, the marginal gains diminish with increasing length.
\vspace{-5px}

\vspace{-5px}
\paragraph{Multi-class versus multi-label classification.}
The challenge of modeling over 1,000 possible event type combinations as a multi-class classification problem is formidable. For example, consider two events, $E_1$ (far\_ad\_bh\_stroke\_DL\_\textit{slice}\_apr\_in) and $E_2$ (far\_ad\_bh\_stroke\_DL\_\textit{drop}\_apr\_in), which differ only in shot technique (\textit{slice} vs. \textit{drop}). Although similar, multi-class classification treats these as distinct classes, thus reducing training efficiency and exacerbating the long-tail distribution bias towards more frequent classes. A more natural approach is multi-label classification, where each event can belong to multiple sub-class elements (e.g., [`far', `ad', `serve', `W', `in']). Thus, $E1$ and $E2$ only differ in shot technique but are identical in other aspects. This adjustment facilitates more effective training and shows an increase in performance, as shown in Table~\ref{tab:ablation}(e). 
% Please see Appendix~\ref{multilabel} for implementation details.
\vspace{-5px}

\vspace{-5px}
\paragraph{Contextual knowledge.} Beyond the statistical results in Table~\ref{tab:results}, analysis of predicted event sequences reveals that current baselines may produce invalid sequences due to logical errors or uncommon practices. For instance, a right-handed player cannot logically direct a forehand shot from the deuce court as ``II'' or ``IO''. Similarly, an event ending in a winner or error should logically conclude the sequence. Additionally, it is uncommon for a player to hit with backhand when the ball is played to their forehand side. Further examples are detailed in Appendix~\ref{error_sequences}. These observations indicate that existing baselines fail to effectively capture event-wise contextual correlations. By adding the CTX module, the performance further increases as shown in Table~\ref{tab:ablation}(f). {\color{black} We also compared BiGRU and Transformer Encoder for the CTX module. BiGRU performed slightly better, likely due to its efficiency in modeling short \emph{event sequences} (usually $< 20$ per clip) with fewer parameters.} 
% It is also helpful to capture contextual knowledge among \fff{} event sequences in other domains.
\vspace{-8px}

\subsection{Generalizability to ``semi-\fff{}'' data}
\vspace{-5px}
\fff{} task possesses broad applicability across numerous real-world domains, such as sports, autonomous driving, surveillance, and production line inspection.
% computer vision in autonomous driving, and advanced manufacturing. 
Nevertheless, creating such a \fff{} dataset necessitates substantial expertise and extensive labeling efforts. 
% We have observed that existing video datasets typically cannot comprehensively cover all three facets of the \fff{} task. In this section, we conducted experiments on some existing ``semi-\fff{}'' datasets (address \emph{partial} aspects of ``fast'', ``frequent'', and ``fine-grained''), including one racket sports Shuttleset (badminton) [\citenum{wang2023shuttleset}], two individual sports FineDiving (diving) [\citenum{xu2022finediving}] and FineGym (gymnastic) [\citenum{shao2020finegym}], and one industrial inspection CCTV-Pipe (pipe defect detection) [\citenum{liu2022videopipe}]. 
We have found that existing video datasets often fail to fully address all three dimensions of the \fff{} task—``fast'', ``frequent'', and ``fine-grained''. In this section, we conducted experiments on several ``semi-\fff{}'' datasets that partially meet these criteria, including Shuttleset [\citenum{wang2023shuttleset}] for badminton (racket sport), FineDiving [\citenum{xu2022finediving}] for diving (individual sports), FineGym [\citenum{shao2020finegym}] for gymnastics (individual sports), {\color{black}SoccerNetV2 [\citenum{mkhallati2023soccernet}] (team sports)}, and CCTV-Pipe [\citenum{liu2022videopipe}] for pipe defect detection (industrial application).
%The main purpose is to show that our method generalizes to other applications as well.
% \begin{itemize}
%     \item \emph{FineDiving} [\citenum{xu2022finediving}] is a diving dataset that contains 29 different action classes (e.g., somersaults twists, and entry) with temporal segment annotations. We detect the starting moments of each action and the action types;
%     \item \emph{FineGym} [\citenum{shao2020finegym}] is a gymnastic dataset with 32 spotting classes, derived from a hierarchy of action types (e.g., balance beam dismounts; floor exercise turns). We identify the temporal boundaries and category of each action;
%     \item \emph{ShuttleSet} [\citenum{wang2023shuttleset}] is a badminton dataset that includes 18 different stroke techniques and the player (near- or far-end) executing the stroke, bringing the total event types to 36 with two sub-classes. We detect hitting moments and stroke event types. 
% \end{itemize}
We report only the F$1_{evt}$ and Edit score, as not all datasets necessitate multi-label classification given their limited event types. For the video encoder, we chose TSM, which consistently outperforms the others on average.

% \zhe{explain why don't compare with TSN and SlowFast methods.}

% EXPLAIN THE LOW PERFORMANCE FOR CCTV-PIPE!!!

% Due to differences in task difficulties and dataset scales, the performances across different domains are very various (some are high and some are low). Take CCTV-Pipe as an example, it focuses on a challenging real-world application of temporal defect localization in urban pipe systems. However, the performance is less optimal than other datasets due to challenges such as ambiguous single-frame temporal annotation for each defect, several defects appear at the same temporal location, severe long-tailed distribution of defect types, and limited dataset size, which are also mentioned and faced in their original paper [\citenum{liu2022videopipe}]. However, on general, as shown in
% Table~\ref{tab:general}, the methods that can handle \dataset{} well tends to perform well in other applications. Furthermore, our \fest{} still outperforms existing baselines across all datasets, showing a good generalizability of detecting \fff{} events in various domains. Therefore, our findings in this paper can also be generalized and benefit other \fff{} related applications.
\vspace{-2px}
Performance across different domains can vary significantly depending on the difficulty of tasks and the scale of datasets. For instance, the CCTV-Pipe dataset, targeting temporal defect localization in urban pipe systems, shows suboptimal performance due to factors such as ambiguous single-frame annotations for each defect, multiple defects at the same time, long-tailed distribution of defect types, and limited dataset size. Our performance is better than the results reported in [\citenum{liu2022videopipe}]. Generally, methods that effectively handle \dataset{} tend to perform well across other applications, as indicated in Table~\ref{tab:general}.  Our \fest{} outperforms existing baselines in all datasets, demonstrating its robust generalizability for detecting ``semi-\fff{}'' events across various domains. {\color{black} While \fff{} event detection benefits from accurate event localization, a high-performing LCL module is not a hard prerequisite (see Appendix~\ref{impact_lcl}).} Therefore, our method can be generalized and benefit broader applications.
\vspace{-10px}

\begin{table*}
\caption{
Experimental results on other ``semi-\fff{}'' datasets.
% Results on \dataset{} with 3 levels of granularity and another challenging badminton dataset ShuttleSet. 
% The best and second-best results are shown in bold and underlined, respectively. 
}
\vspace{-2mm}
\centering
\small
\resizebox{\textwidth}{!}{
\begin{tabular}{@{}llcccccccccccc@{}}
\toprule
 & \multicolumn{2}{c}{ShuttleSet [\citenum{wang2023shuttleset}]} & \multicolumn{2}{c}{FineDiving [\citenum{xu2022finediving}]} & \multicolumn{2}{c}{FineGym [\citenum{shao2020finegym}]} &\multicolumn{2}{c}{\color{black}SoccerNetV2 [\citenum{deliege2021soccernet}]} & \multicolumn{2}{c}{CCTV-Pipe [\citenum{liu2022videopipe}]} \\
\cmidrule(r){2-3} \cmidrule(lr){4-5} \cmidrule(l){6-7} \cmidrule(l){8-9} \cmidrule(l){10-11} 
 Head arch. & F$1_{evt}$ & Edit & F$1_{evt}$ & Edit & F$1_{evt}$ & Edit & {\color{black}F$1_{evt}$} & {\color{black}Edit} & F$1_{evt}$ & Edit \\
\midrule
MS-TCN [\citenum{farha2019ms}]
% \cite{lin2022swinbert} 
& 70.3 & 74.4 & 65.7 & 92.2 & 57.6 & 65.3 &{\color{black}43.4} & {\color{black}74.5} &25.8 &31.3\\
ASformer [\citenum{yi2021asformer}]
% \cite{lin2022swinbert} 
& 55.9 & 70.6 & 49.9 & 87.6 & 53.6 & 66.3 &{\color{black}46.3} & {\color{black}76.1} &15.4 &33.4\\
% \midrule
% \multicolumn{10}{l}{\textit{(b) Temporal Action Localization \& Spotting}} \vspace{1mm} \\
G-TAD [\citenum{xu2020g}]
% \cite{behrmann2022unified} 
& 48.2 & 61.1 & 52.1 & 82.6 & 45.8 & 51.4 &{\color{black}42.3} & {\color{black}72.3} & 31.3 & 33.6 \\
ActionFormer [\citenum{zhang2022actionformer}]
% \cite{lin2022swinbert} 
& 62.1 & 67.5 & 68.3 & 92.4 & 54.0 & 59.7 &{\color{black}43.0} & {\color{black}64.6} &18.8 &29.5 \\
E2E-Spot [\citenum{hong2022spotting}]
% \cite{hong2022spotting} 
& 70.2 & 75.0 & 75.8 & 93.7 & 62.1 & 65.4 &{\color{black}46.2} & {\color{black}72.9} &27.2 &35.2 \\
\midrule
\fest{} 
% \cite{hong2022spotting} 
& \textbf{70.7} & \textbf{77.1} & \textbf{77.6} & \textbf{95.1} & \textbf{70.9} & \textbf{70.7} &\textbf{\color{black}48.1} & \textbf{\color{black}76.6} & \textbf{37.0} &\textbf{39.5} \\
\bottomrule
\end{tabular}
}
\vspace{-15px}
\label{tab:general}
\end{table*}

\section{Conclusion and Future Work}
\vspace{-7px}
In this study, we addressed the challenge of analyzing fast, frequent, and fine-grained (\fff{}) events from videos by introducing \dataset{}, a benchmark for precise temporal \fff{} event detection. 
\dataset{} datasets usually feature detailed event types (approximately 1,000), annotated with precise timestamps, and provide multi-level granularity.
We have also developed a general annotation toolchain that enables domain experts to create \fff{} datasets, thereby facilitating further research in this field. Moreover, we proposed \fest{}, an end-to-end model that effectively detects complex event sequences from videos, using a combination of visual features and contextual sequence refinement. Our comprehensive evaluations and ablation studies of leading methods in temporal action understanding on \dataset{} highlighted their performance and provided critical insights into their capabilities and limitations. Moving forward, we aim to extend the scope of \fff{} task to more real-world scenarios and advance the development of \fff{} video understanding.

\clearpage
\subsubsection*{Acknowledgments}
This research is supported by the AI Singapore (AISG3-RP-2022-030).  Any opinions, findings and conclusions or recommendations expressed in this material are those of the author(s) and do not reflect the views of funding bodies.

\setcitestyle{numbers}
\bibliography{iclr2025_conference}
\bibliographystyle{iclr2025_conference}

\clearpage
\appendix
% Our \dataset{} dataset and the benchmark code are accessible at \url{https://github.com/F3EST/F3Tennis}

% \section{License} \label{license}
% The \dataset{} dataset is under the CC BY 4.0 International license. Please refer to \url{https://creativecommons.org/licenses/by/4.0/legalcode} for license detail.

\section{Preliminary Results for Multi-Modal LLMs}
\label{gpt}
\begin{figure*}[t]
\centering
    \includegraphics[width=\linewidth]{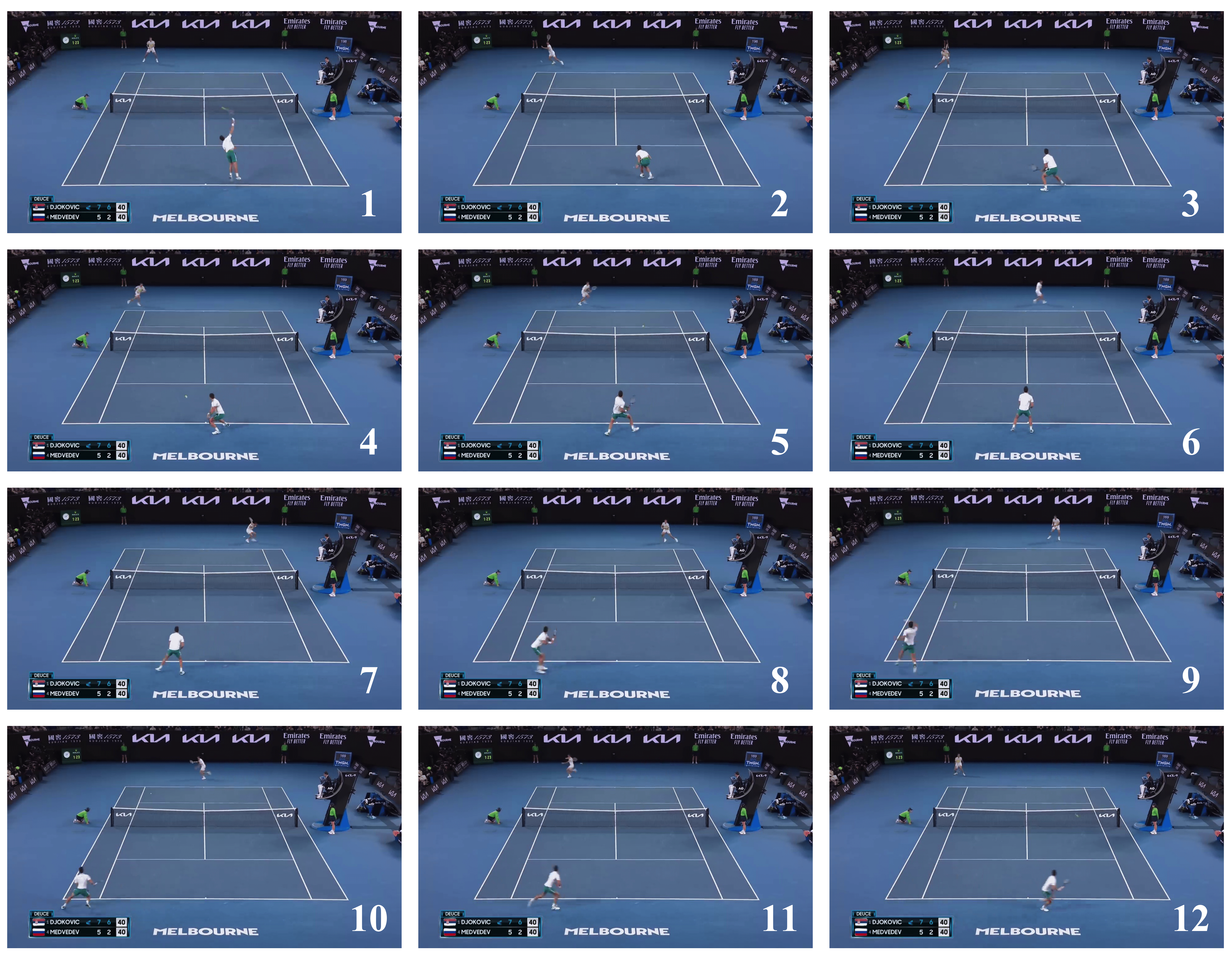}
    % \vspace{-10px}
\caption{Video frames from a tennis rally.}
\label{fig:gpt4}
\end{figure*}

We conducted experiments using GPT4 \cite{openai2023gpt4}, specifically employing the GPT4-vision-preview feature. As depicted in Figure~\ref{fig:gpt4}, we input a sequence of 12 consecutive frames from a tennis rally into the system. This sequence encompasses a total of 6 shots: 
\begin{itemize}
    \item wide serve by the near player,
    \item forehand return down the middle by the far player
    \item backhand stroke cross-court by the near player
    \item backhand stroke cross-court by the far player
    \item forehand stroke inside-in by the near player
    \item forehand stroke cross-court by the far player (last shot).
\end{itemize}

Our initial tests are aimed at assessing GPT4-vision-preview's ability to decipher basic contextual information. As demonstrated in Table~\ref{tab:QA1}, GPT4 demonstrates proficiency in recognizing the type of sport, deciphering information from the scoreboard, and identifying the tournament category. However, it exhibits some inaccuracies in determining the exact game points.

\begin{table}[h]
\caption{User Prompt and GPT4's Response for Question 1. The text highlighted in bold represents correct responses, underlined text indicates incorrect responses, and text in red within brackets is the correction.}
\centering
{\small % Reducing the font size
\begin{tabular}{p{0.95\columnwidth}}
\toprule
\textbf{User} \\
These are images from a video. What information can you get? \\ 
\midrule
\textbf{GPT4} \\
The images you provided are \textbf{sequential frames from a tennis match}. Based on the information visible on the scoreboard in the images, it seems to be an intense moment with the following details: 
\begin{itemize}[leftmargin=*]
    \item The \textbf{score is at deuce}, indicating that both players have won the same number of points after each having scored \textbf{at least three points in this particular game}.
    \item The player labeled \textbf{`Djokovic' is leading the match}, having \underline{won the first set 7-6} {\color{red}(7-5)} and is ahead in the second set with a \underline{score of 4-2} {\color{red}(6-2)}.
    \item The location is marked as \textbf{Melbourne}, which suggests that this match might be \textbf{taking place during the Australian Open}, a major tennis tournament that is held annually in Melbourne.
    \item The players are \textbf{engaged in a rally, with one player serving and the other returning the ball}.
\end{itemize}
The images show a snapshot of the action, \textbf{capturing the movement and intensity of professional tennis play}.\\
\bottomrule
\end{tabular}
}
\label{tab:QA1}
\end{table}

However, when we shift our focus to more complex queries requiring deeper insight, as shown in Table~\ref{tab:QA2}, we observe a notable disparity. Although GPT4 generates coherent responses, its precision in recognizing and interpreting the fine-grained sequence of actions within the rally is limited. As there are too many incorrect responses, we do not use GPT4 in the experiment for comparison.

\begin{table}[t!]
\caption{User Prompt and GPT4's Response for Question 2. The text highlighted in bold represents correct responses, underlined text indicates incorrect responses, and text in red within brackets is the correction.}
\centering
{\small % Reducing the font size
\begin{tabular}{p{0.95\columnwidth}}
\toprule
\textbf{User} \\
These images show the video of a whole tennis rally between two right-hand players. Describe what's likely going on in each frame. Try to find connections across frames and describe the shot type (serve, forehand, backhand) and direction (T, wide, body, cross-court, down the line, down the middle, inside-in, inside-out). \\ 
\midrule
\textbf{GPT4} \\
% 1. Player A executes a \textbf{serve}, likely aiming \textbf{wide} towards the deuce side of the service box.\\
% 2. Player B \textbf{returns} the serve, possibly with a \underline{backhand} {\color{red}(forehand)} stroke aimed \underline{cross-court} {\color{red}(down the middle)}.\\
% 3. Player A \textbf{moves} to hit the ball, \textbf{preparing} for what might be a \underline{forehand} {\color{red}(backhand)} shot.\\
% 4. Player A hits the \underline{forehand} {\color{red}(backhand)}, directing it \textbf{cross-court} towards Player B's backhand side.\\
% 5. Player B \textbf{runs to reach the ball}, setting up for a \textbf{backhand} return.\\
% 6. Player B hits the ball with a \textbf{backhand stroke}, likely aiming back \textbf{cross-court}.\\
% 7. Player A anticipates the ball's trajectory, \textbf{moving to make a return shot}.\\
% 8. Player A approaches the ball, preparing to hit it, potentially with a \textbf{forehand} stroke.\\
% 9. Player A contacts the ball with a \textbf{forehand}, aiming the shot potentially \underline{down the line or cross-court} {\color{red}(inside-in)}.\\
% 10. Player B reacts \textbf{quickly to chase the ball down}, possibly to hit a \textbf{forehand}.\\
% 11. Player B executes the \textbf{forehand} shot, aiming it across the court.\\
% 12. Player A tracks the ball's path, preparing to return.\\

1. One player has just \textbf{served} the ball. The serve appears to be going towards the \underline{T (center)} {\color{red}(wide)} of the service box on the opponent's side.\\
2. The receiving player is getting ready to \underline{return the serve with a backhand stroke} {\color{red}(forehand)}, suggesting the serve was directed to his backhand side.\\
3. The receiver has hit the ball back, likely aiming \underline{cross-court} {\color{red}(down the middle)} based on their body position and the angle of the racket at impact.\\
4. The server has \textbf{moved towards the center} of the baseline in preparation to return the shot, suggesting they \textbf{anticipated a cross-court return}.\\
5. The server hits the ball, potentially with a \underline{forehand} {\color{red}(backhand)} stroke, aiming the shot either \underline{cross-court or down the line} {\color{red}(cross-court)}, as they are positioned on the deuce side of the court.\\
6. The \textbf{ball is in transit}, and the players are positioning themselves for the next shot. The \textbf{receiver is moving to his left}, probably anticipating a ball to his \textbf{forehand} side.\\
7. The receiver is setting up for a \textbf{forehand return}, indicating the previous shot was indeed to his forehand.\\
8. The ball is struck, with the player's body and racket angle indicating a possible \textbf{cross-court} shot.\\
9. The original server is on the run to the opposite side, preparing for a possible \underline{backhand return} {\color{red}(forehand)}, which suggests the previous shot was a cross-court \underline{forehand} {\color{red}(backhand)}.\\
10. The player makes contact with a \underline{backhand stroke} {\color{red}(forehand)}, likely sending the ball either \underline{down the line or cross-court} {\color{red}(cross-court)} given their body orientation.\\
11. The players reposition again for the next shot, with the server \textbf{transitioning toward the center}, preparing for either a forehand or backhand.\\
12. The \underline{server prepares for a backhand return} {\color{red}(forehand)}, suggesting the ball was returned to his backhand side.\\

\bottomrule
\end{tabular}
}
\label{tab:QA2}
\end{table}

\section{Tennis Lexicon.} \label{lex}
\begin{figure}[ht!]
\centering
    \includegraphics[width=0.7\linewidth]{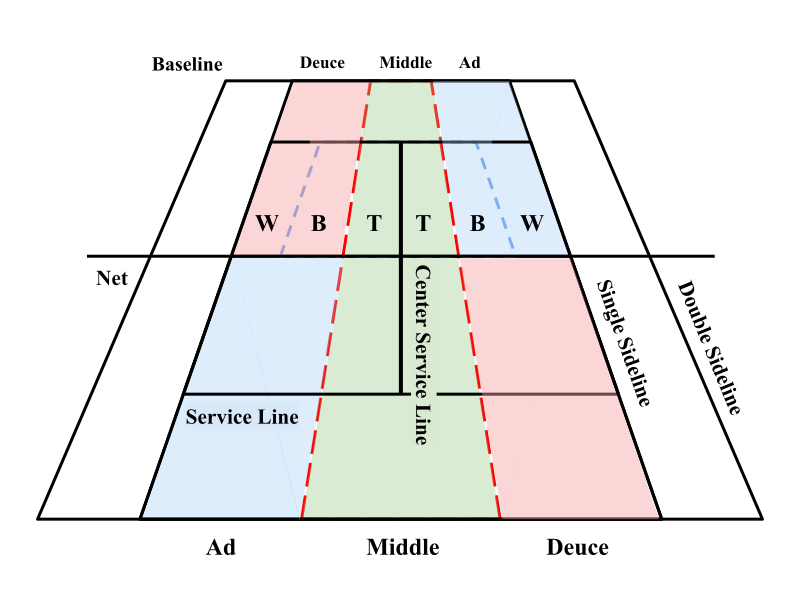}
    % \vspace{-10px}
\caption{Tennis court and related terminologies.}
\label{fig:tennis_court}
\end{figure}

In this section, we will provide an overview of technical tennis terms used in our model based on official terminology from the USTA\footnote{\url{https://www.usta.com/en/home/improve/tips-and-instruction.html}}. A standard tennis court is depicted in Fig.~\ref{fig:tennis_court}. Each side of the court can be divided into three regions: the deuce court (red), the middle court (green), and the ad court (blue). Additionally, the service box is defined by the boundaries of the net, service line, center service line, and single sideline. It includes three sub-regions, the \textit{T}, \textit{B}, and \textit{W} areas.

\label{subsec:tennis-terms}

In tennis, the initial shot of a point is called a \textit{serve}. The ball must land in the service box and is usually hit overhead. On odd points, the server serves to the receiver's right-hand side, named \textit{deuce court serve}; while on even points, the server serves to the receiver's left-hand side, named \textit{ad court serve}. The server can aim to hit the ball in the \textit{T}, \textit{B}, or \textit{W} area of the service box.

The shot taken by the receiver after a serve is called a \textit{return}. If it lands in bounds after crossing the net. Subsequent shots are referred to as \textit{strokes}. Players can hit the ball cross-court, down the line, down the middle, inside-in or inside-out using either their \textit{forehand} ({fh}) or \textit{backhand} ({bh}). \textit{Cross-court} ({CC}) means a shot that travels diagonally from the player's position (e.g., a right-hand player's forehand shot from his deuce/middle court to the opponent's deuce court). \textit{Down the line} ({DL}) refers to a straight shot from their position (e.g., a right-hand player's backhand shot from his ad court to the opponent's deuce court). \textit{Down the middle} ({DM}) means a shot toward the opponent's middle court. \textit{Inside-out} ({IO}) / \textit{inside-in} ({II}) refers to a player changing the shot from backhand to forehand or vice versa. For example, if a right-hand player hits a forehand shot from his ad/middle court to the opponent's ad court, this is called a forehand inside-out. Similarly, if a right-hand player hits a forehand shot from his ad court to the opponent's deuce court, this is called a forehand inside-in.

Generally, a player can approach the net on a ball that lands around the service line or shorter, or if they recognize that their opponent is out of position and is likely to provide a weak ball. This is called an \textit{approach} (apr) shot, which is defined as an offensive shot that allows a tennis player to transition from the baseline to the net, hitting a forehand or a backhand. If a player choose to stay at the baseline or already positioned at the net, we call it \textit{-} (n-apr). Players can also apply different shot techniques under certain conditions. A \textit{ground stroke} (gs) is a basic tennis shot executed after the ball bounces once on the court, typically used from the baseline. A \textit{slice} is a shot where the player imparts a backspin on the ball, causing it to travel slowly and with a lower trajectory, which can disrupt an opponent's timing. A \textit{volley} is hit before the ball bounces, usually performed near the net to shorten the point. A \textit{lob} is a shot that sends the ball high and deep into the opponent's court, often used to counter opponents who are close to the net.

Each shot has four possible outcomes: \textit{in-bound (in)}, where the ball lands within the opponent's court boundaries; \textit{winner (win)}, a shot that is successfully placed where the opponent cannot return it, directly winning the point; \textit{forced error (fe)}, where the shot is so challenging that the opponent makes an error trying to return it, often due to the pressure exerted by the aggressive play; and \textit{unforced error (ufe)}, where the player fails to return the ball in the court without external pressure, typically due to a mistake in execution.

\section{\fff{} Events in Other Domains} \label{other-fff-events}
For the badminton dataset, each event consists of 6 \emph{sub-classes}, denoted as $sc_1, sc_2, ..., sc_6$: 
\begin{itemize}[noitemsep,leftmargin=15pt]
\item $sc_1$ -- \textit{hit by which player}: (1) near- or (2) far-end player; 
\item $sc_2$ -- \textit{hit from which court location}: (3) left, (4) middle, or (5) right court; 
\item $sc_3$ -- \textit{hit at which side of the body}: (6) forehand or (7) backhand; 
\item $sc_4$ -- \textit{shot type}: (8) serve-short, (9) serve-long, (10) net, (11) smash, (12) lob, (13) clear, (14) drive, (15) drop, (16) push or (17) rush; 
\item $sc_5$ -- \textit{shot direction}: (18) T, (19) B, (20) W, (21) CC, (22) DL, or (23) DM; 
\item $sc_6$ -- \textit{shot outcome}: (24) in, (25) winner, or (26) error.
\end{itemize}
Altogether, there are 26 \emph{elements} and 1,008 \emph{event types} based on various combinations.

For the table tennis dataset, each event consists of 7 \emph{sub-classes}, denoted as $sc_1, sc_2, ..., sc_7$: 
\begin{itemize}[noitemsep,leftmargin=15pt]
\item $sc_1$ -- \textit{hit by which player}: (1) near- or (2) far-end player; 
\item $sc_2$ -- \textit{hit from which court location}: (3) left, (4) middle, or (5) right court; 
\item $sc_3$ -- \textit{hit at which side of the body}: (6) forehand or (7) backhand; 
\item $sc_4$ -- \textit{shot spin}: (8) top, (9) bottom, or (10) side; 
\item $sc_5$ -- \textit{shot type}: (11) serve, (12) push, (13) chop, (14) drive, (15) block, or (16) smash;
\item $sc_6$ -- \textit{shot direction}: (17) straight-long, (18) straight-short, (19) diagonal-long, or (20) diagonal-short;
\item $sc_7$ -- \textit{shot outcome}: (21) in, (22) winner, or (23) error.
\end{itemize}
Altogether, there are 23 \emph{elements} and 1,296 \emph{event types} based on various combinations.

For the tennis double dataset, each event consists of 7 \emph{sub-classes}, denoted as $sc_1, sc_2, ..., sc_6$: 
\begin{itemize}[noitemsep,leftmargin=15pt]
\item $sc_1$ -- \textit{hit by which player}: (1) near- or (2) far-end player; 
\item $sc_2$ -- \textit{hit from which court location}: (3) deuce or (4) ad court; 
\item $sc_3$ -- \textit{hit at which side of the body}: (5) forehand or (6) backhand; 
\item $sc_4$ -- \textit{shot type}: (7) serve, (8) return, (9) volley, (10) lob, (11) smash, or (12) swing;
\item $sc_6$ -- \textit{shot direction}: (13) T, (14) B, (15) W, (16) CC, (17) DL, (18) II, or (19) IO; 
\item $sc_7$ -- \textit{serving formation}: (20) conventional, (21) I-formation, (22) Australian, or (23) non-serve.
\item $sc_7$ -- \textit{shot outcome}: (24) in, (25) winner, or (26) error.
\end{itemize}
Altogether, there are 26 \emph{elements} and 744 \emph{event types} based on various combinations.

For the full elements and event types, please refer to \url{https://github.com/F3SET/F3Set/blob/main/data/}.

\begin{table}[t]
\caption{Distribution of elements across sub-classes in the \dataset{}.}
\vspace{1mm}
\centering
\small
\begin{tabular}{@{}llrcccccccccccccc@{}}
\toprule
 Sub-Class & Element & Count & Proportion (\%) in Sub-Class \\
\midrule
\multirow{2}{*}{$sc_1$} & \textit{near} & 21,467 & 50.1\% \\
                        & \textit{far} & 21,362 & 49.9\%\\ 
\midrule
\multirow{3}{*}{$sc_2$} & \textit{deuce} & 14,474 & 33.8\% \\
                        & \textit{ad} & 16,310 & 38.1\%\\
                        & \textit{middle} & 12,045 & 28.1\%\\
\midrule
\multirow{2}{*}{$sc_3$} & \textit{forehand} & 27,802 & 64.9\%\\
                        & \textit{backhand} & 15,027 & 35.1\%\\ 
\midrule
\multirow{3}{*}{$sc_4$} & \textit{serve} & 11,584 & 27.0\% \\
                        & \textit{return} & 8,216 & 19.2\% \\
                        & \textit{stroke} & 23,029 & 53.8\% \\
\midrule
\multirow{8}{*}{$sc_5$} & \textit{T} & 4,428 & 10.3\%\\
                        & \textit{Body} & 2,241 &  5.2\%\\
                        & \textit{Wide} & 4,915 &  11.5\%\\
                        & \textit{cross-court} & 11,933 & 27.9\% \\
                        & \textit{down the line} & 3,521 & 8.2\% \\
                        & \textit{down the middle} & 11,040 & 25.8\% \\
                        & \textit{inside-in} & 608 & 1.4\% \\
                        & \textit{inside-out} & 4,143 & 9.7\% \\
\midrule
\multirow{6}{*}{$sc_6$} & \textit{ground stroke} & 38,287 & 89.4\% \\
                        & \textit{slice} & 3,358 & 7.8\% \\
                        & \textit{volley} & 497 & 1.2\% \\
                        & \textit{lob} & 334 & 0.8\% \\
                        & \textit{drop} & 236 & 0.5\% \\
                        & \textit{smash} & 117 & 0.3\% \\
\midrule
\multirow{2}{*}{$sc_7$} & \textit{approach} & 964 & 2.3\% \\
                        & \textit{non-approach} & 41,865 & 97.7\% \\
\midrule
\multirow{4}{*}{$sc_8$} & \textit{in-bound} & 31,245 & 73.0\% \\
                        & \textit{winner} & 3,734 & 8.7\% \\
                        & \textit{forced error} & 2,808 & 6.5\% \\
                        & \textit{unforced error} & 5,042 & 11.8\% \\
\bottomrule
\end{tabular}
% }
% \vspace{-3px}
\label{tab:distribution}
\end{table}
\normalsize

\section{Data Statistics} \label{stats}
{\color{black} We recognize the importance of testing generalizability under diverse real-world conditions, including variations in camera angles, court types, weather, and illumination. While professional competition videos are filmed under relatively standardized conditions, we have ensured that \dataset{} captures a significant level of diversity in these factors:

\begin{itemize}[leftmargin=*]
    \item Camera Angles: The dataset includes videos from 114 broadcast matches across various tournaments, each exhibiting different camera angles. Additionally, individual matches often feature multiple perspectives: standard bird’s-eye view and low-angle view shots from behind one player, where the foreground player appears significantly larger than the background player.
    \item Court Types: The dataset covers all three tennis court surfaces—hard court, clay court, and grass court—with diverse color schemes (e.g. blue, green, red, black, green, purple, etc).
    \item Weather Conditions and Illumination: The videos in \dataset{} reflect diverse weather and lighting scenarios, including day and night matches, cloudy weather, indoor and outdoor games, and challenging conditions such as partial sunlight casting shadows on the court. For outdoor matches, some videos feature strong sunlight on parts of the court, making it harder to track the ball or players in those areas. Indoor matches also vary in brightness, with some tournaments having brighter lighting setups and others being relatively dimmer.
\end{itemize}

To further enhance the diversity, we have collected videos from college-level matches for tennis doubles. We are also actively expanding \dataset{} by incorporating more matches from platforms like UTR\footnote{\url{https://www.utrsports.net/}} and junior-level competitions using our annotation toolchain.}

\begin{table}
\small
\centering
\caption{Summary of \dataset{} badminton dataset statistics.}
\vspace{-1.5mm}
{\color{black}
\begin{tabular}{lc}
\toprule
{Category}        & {Details}                     \\ 
\midrule
Matches                 & 10 broadcast matches                \\
Players                 & 16 (16 men)                \\
Handedness              & 13 right-handed, 3 left-handed       \\
Frame Rate              & 25--30 FPS                           \\
Clips                   & 112 rallies                       \\
Average Clip Duration   & 13.5 seconds                          \\
Total Shots             & 1692                               \\
Shots Per Rally         & 1 to 62                              \\ 
\bottomrule
\end{tabular}
}
\vspace{-10px}
\label{tab:badminton_stats}
\end{table}
\normalsize

{\color{black}\subsection{Additional \fest{} statistics}
Key statistics for $\dataset{}$ badminton dataset are summarized in Table~\ref{tab:badminton_stats}; for $\dataset{}$ table tennis dataset are summarized in Table~\ref{tab:table-tennis_stats}; for $\dataset{}$ tennis doubles dataset are summarized in Table~\ref{tab:tennis-doubles_stats}.}

\begin{table}
\small
\centering
\caption{Summary of \dataset{} table tennis dataset statistics.}
\vspace{-1.5mm}
{\color{black}
\begin{tabular}{lc}
\toprule
{Category}        & {Details}                     \\ 
\midrule
Matches                 & 5 broadcast matches                \\
Players                 & 9 (4 men, 5 women)                \\
Handedness              & 9 right-handed       \\
Frame Rate              & 25--30 FPS                           \\
Clips                   & 42 rallies                       \\
Average Clip Duration   & 5.9 seconds                          \\
Total Shots             & 361                               \\
Shots Per Rally         & 2 to 36                              \\ 
\bottomrule
\end{tabular}
}
\vspace{-10px}
\label{tab:table-tennis_stats}
\end{table}
\normalsize

\begin{table}
\small
\centering
\caption{Summary of \dataset{} tennis doubles dataset statistics.}
\vspace{-1.5mm}
{\color{black}
\begin{tabular}{lc}
\toprule
{Category}        & {Details}                     \\ 
\midrule
Matches                 & 8 broadcast matches                \\
Players                 & 24 (24 men)                \\
Handedness              & 24 right-handed       \\
Frame Rate              & 25--30 FPS                           \\
Clips                   & 78 rallies                       \\
Average Clip Duration   & 6.0 seconds                          \\
Total Shots             & 645                               \\
Shots Per Rally         & 2 to 21                              \\ 
\bottomrule
\end{tabular}
}
\vspace{-10px}
\label{tab:tennis-doubles_stats}
\end{table}
\normalsize

Furthermore, we present additional statistics related to the \dataset{} (tennis single) datasets. 
% Figure~\ref{fig:classes} illustrates the annotation structure of \dataset{}, which comprises 8 sub-classes, 30 sub-class elements, and 1,108 distinct event types. 
Table~\ref{tab:distribution} details the frequency and proportional occurrence of elements within each sub-class. For the full elements and event types, please refer to \url{https://github.com/F3SET/F3Set/blob/main/data/f3set-tennis/}.

\section{Ethical Considerations for \dataset{} Datasets} \label{ethical}
The \dataset{} is constructed from publicly available video data sourced from YouTube, specifically from officially broadcasted tennis tournaments featuring professional players. This document outlines the ethical considerations related to data collection, copyright compliance, privacy concerns, and bias mitigation.

The dataset consists of publicly available sports broadcasts, ensuring compliance with ethical and legal standards. We do not store or distribute local copies of the videos unless explicitly permitted under Creative Commons or similar licenses. If a video is removed or becomes unavailable, we update our dataset accordingly while ensuring adherence to copyright policies. To comply with YouTube’s Terms of Service, we provide only video URLs, ensuring that the content remains under the control of rights holders.

Our dataset exclusively features professional players in widely broadcasted tournaments. As such, the dataset includes individuals who are already in the public domain through official broadcasts. No private or off-court data is collected, and annotations focus solely on event-based information. The dataset is strictly intended for research purposes, and users must ensure ethical compliance in their applications. A disclaimer is provided, explicitly stating that the dataset should not be used beyond academic research.

We take steps to mitigate potential biases in the dataset. The dataset does not incorporate or filter data based on nationality or ethnicity, ensuring a broad and representative scope. We encourage users to evaluate and report any potential biases that may emerge in model training. Additionally, the dataset will be periodically reviewed and updated based on community feedback to ensure fairness.

The \dataset{} dataset has been designed to align with ethical standards, ensuring responsible use of publicly available content. We have addressed concerns regarding copyright, privacy, and bias. By emphasizing academic-only use and compliance with licensing terms, we provide a valuable resource for sports analytics research while respecting the rights and privacy of all involved stakeholders.

\section{Implementation Details} \label{implement}
\subsection{Baseline models}
\begin{itemize}[leftmargin=*]
    \item \textit{TSN} [\citenum{wang2018temporal}] utilizes a purely 2D CNN architecture. Each frame is processed independently with RGB images as inputs, employing the RegNet-Y 200MF architecture [\citenum{radosavovic2020designing}] as the backbone.
    \item \textit{TSM} [\citenum{lin2019tsm}] incorporates a temporal shift mechanism within the 2D convolutional process of video encoders. This shift along the temporal axis mimics a cost-free 1D convolution, enabling efficient extraction of spatiotemporal features through subsequent convolutions on shifted inputs. The RegNet-Y 200MF backbone is augmented with Temporal Shift Modules (TSM) [\citenum{lin2019tsm}], integrated at strategic points within each residual block, specifically targeting a quarter of the channels, adjusted to the nearest multiple of four.
    \item \textit{SlowFast} [\citenum{feichtenhofer2019slowfast}] features dual pathways: the slow pathway processes frames at a sparse rate to capture high-level information, and the fast pathway processes at a higher frame rate with fewer channels to capture detailed motion information efficiently. These pathways are integrated at various stages to enhance the assimilation of motion information. We utilize the SlowFast network as our video encoder, specifically the ``SlowFast 4$\times$16, R50'' variant. This model samples $N$ and $N/8$ frames in their fast and slow pathways, respectively, which are then resized to length $N$ and concatenated.
\end{itemize}

The selection of RegNet-Y 200MF over traditional ResNet models [\citenum{he2016deep}] is based on its recent advancements, lower parameter count (3.2M compared to ResNet-18's 11.7M), and superior performance in image classification benchmarks [\citenum{deng2009imagenet}]. This architecture allows flexibility for integrating alternative 2D CNN designs. Beginning with pretrained ImageNet-1K weights [\citenum{deng2009imagenet}], the encoder is meticulously fine-tuned to our specific dataset needs.

% We implement and train the baseline models on \dataset{} dataset in an end-to-end manner. The video encoder takes video clip $X$ down-scaled and cropped to 224 $\times$ 224 to extract frame-wise visual features $\textbf{F}_{vid} \in \mathbb{R}^{N \times d}$. Subsequently, each head module processes $\textbf{F}_{vid}$ to produce encoded features $\textbf{F}_{enc} \in \mathbb{R}^{N \times d'}$. To identify hitting moments and their respective event types, frame-wise dense classification is applied on $\textbf{F}_{enc}$, resulting in predictions for each frame $P=(p_1, \ldots, p_N)$, where $p_i$ is classified as either background or one of the event types. 
% % Given $J$ distinct action classes, denoted as $C = \{c_1, ..., c_J\}$, $p_i \in \{c_{background}\} \cup C$. 
% The cross-entropy loss is computed as
% \begin{equation}
%     L = \frac{1}{N} \sum^{N}_{i=1} \text{CrossEntropy}(p_i, y_i),
% \end{equation}
% where $y_i$ represents the ground truth for frame $i$.

Let $J$ denotes the number of event types.

\begin{itemize}[leftmargin=*]
    \item \textit{MS-TCN} [\citenum{farha2019ms}] employs successive layers of dilated convolutions to capture long-range temporal dependencies in sequence modeling tasks. We adapt the code from [\citenum{farha2019ms}], using dilated temporal convolution networks. We use 3 TCN stages for our MS-TCN baselines and a depth of 5 layers for each stage. Each layer has dimension of 256. Per-frame predictions are made with a fully connected layer that maps from 256 to $J + 1$, where each frame is classified as either background or one of the event types.
    \item \textit{ASFormer} [\citenum{yi2021asformer}] leverages a transformer-based architecture with segment embeddings to enhance temporal action segmentation. We use code and settings from the implementation by [\citenum{yi2021asformer}].
    \item \textit{G-TAD} [\citenum{xu2020g}] employs a graph convolutional network to model complex temporal relationships between video segments, enhancing the accuracy of action detection. We use the GCNeXt block architecture proposed by [\citenum{xu2020g}], which produces a dimensional feature encoding $H$ of 384, 384, and 768 for TSN, TSM, and SlowFast, respectively, for each frame. Per-frame predictions are made with a fully connected layer mapping from $H$ to $J + 1$.
    \item \textit{ActionFormer} [\citenum{zhang2022actionformer}] employs transformer networks for efficient single-shot temporal action localization, using multiscale features and local self-attention. We employ the architecture proposed by [\citenum{zhang2022actionformer}] and produce a dimensional feature encoding $H$ of 384, 384, and 768 for TSN, TSM, and SlowFast, respectively. Per-frame predictions are made with a fully connected layer mapping from $H$ to $J + 1$.
    \item \textit{E2E-Spot} [\citenum{hong2022spotting}] utilizes a bidirectional GRU layer to facilitate long-term temporal reasoning for precise action spotting. We use a 1-layer bidirectional GRU [\citenum{dey2017gate}] with dimensions $H$ of 384, 384, and 768 for TSN, TSM, and SlowFast, respectively. Per-frame predictions are made with a fully connected layer, from $H$ to $J + 1$.
\end{itemize}

\subsection{\fest{}}
\paragraph{Training.} We train all components of \fest{} in an end-to-end manner. The video encoder, equipped with a 2D CNN {\color{black}(i.e., RegNet-Y 200MF)} and TSM, is initialized using pre-trained ImageNet-1K weights [\citenum{deng2009imagenet}] and subsequently fine-tuned the targeted dataset. The LCL processes frame-wise spatio-temporal features to perform dense predictions, distinguishing event instances from the background. The MLC receives ground truth event instances to concentrate on the classification task, assuming accurate localization. Classification is executed densely, yet losses are computed only on frames that contain event instances. For the CTX, we input the predicted event sequences. The sequence is obtained by combining the outputs from LCL, MLC, and ground truth event locations. We then feed the predicted sequence to CTX to obtain a refined one.  
% , randomly replace $P=30\%$ of these events with the current MLC predictions, and generate refined event sequences that integrate visual observations and causal correlations. Losses are calculated on the replaced event tokens. 
Overall, our composite loss function is defined as $L = L_{LCL} + L_{MLC} + L_{CTX}$.

\paragraph{Inference.} During the inference phase, the MLC uses localization results from the LCL to predict corresponding event types. The CTX processes the event sequence generated from the LCL and MLC outputs, producing a new sequence of the same length that incorporates both visual predictions and contextual correlation across events.

{\color{black} \subsection{Model Implementation Details}

For both baseline models and the proposed \fest{}, the training protocol processes sequences of 96 frames with a stride of 2. Batch size is set to 4. Standard data augmentation techniques, including cropping and color adjustments, are applied during training to enhance data diversity and improve model robustness; these augmentations are omitted during testing. Input frames are resized to a height of 224 pixels, followed by a random crop to a 224 $\times$ 224 square, ensuring preservation of essential visual information by selectively adjusting the width. Techniques such as cropping and color jittering are employed to further augment the dataset and fortify the models against overfitting. 

Each model performs dense, per-frame classification to identify event types and their precise timestamps. Given the imbalance in event distribution, where less than 3\% of frames correspond to specific event instances, the loss weight for foreground classes is increased fivefold relative to background classes to address this disparity.

The models are optimized using the AdamW optimizer, with a learning rate schedule controlled via cosine annealing. Training is conducted over 50 epochs, with each epoch taking approximately 10 minutes on an RTX 4090 GPU. The initial learning rate is set to 0.001, with three linear warm-up steps before transitioning to a cosine decay schedule. For computationally intensive video encoders, such as SlowFast and VTN, a smaller initial learning rate of 0.0001 is used to ensure stable convergence.}

{\color{black} \section{Additional Experimental Results} \label{results_sup}
In this section, we provide the full experimental result table as shown in Table~\ref{tab:results_full}.}

\begin{table}
\caption{
Full experimental results on \dataset{} (tennis) with 3 levels of granularity.
% The best and second-best results are shown in bold and underlined, respectively. 
} 
\vspace{-1mm}
\centering
\small
\resizebox{\textwidth}{!}{
\begin{tabular}{@{}llccccccccccc@{}}
\toprule
 & & \multicolumn{3}{c}{$\dataset{}\,(G_{high})$} & \multicolumn{3}{c}{$\dataset{}\,(G_{mid})$} & \multicolumn{3}{c}{$\dataset{}\,(G_{low})$} \\
\cmidrule(r){3-5} \cmidrule(lr){6-8} \cmidrule(l){9-11}
Video encoder & Head arch. & F$1_{evt}$ & F$1_{elm}$ & Edit & F$1_{evt}$ & F$1_{elm}$ & Edit & F$1_{evt}$ & F$1_{elm}$ & Edit \\
\midrule
TSN [\citenum{wang2018temporal}]
&MS-TCN [\citenum{farha2019ms}]
&15.9 & 59.8 & 53.5 &23.2 & 60.9 & 65.8 &45.7 & 70.4 & 72.8 \\
&ASformer  [\citenum{yi2021asformer}]
&11.9 & 54.3 & 49.8 &17.3 & 56.1 & 62.5 &40.3 & 67.3 & 70.3 \\
&G-TAD [\citenum{xu2020g}]
&6.0 & 47.5 & 24.7 &14.1 & 52.1 & 48.6 &19.9 & 57.4 & 44.7 \\
&ActionFormer [\citenum{zhang2022actionformer}]
&18.4 & 60.6 & 55.2 &24.8 & 61.9 & 67.3 &48.7 & 70.6 & 72.2 \\
&E2E-Spot [\citenum{hong2022spotting}]
&24.7 & 65.3 & 60.1 &31.5 & 66.2 & 71.0 &53.5 & 73.6 & 75.0 \\
\midrule
SlowFast [\citenum{feichtenhofer2019slowfast}]
&MS-TCN [\citenum{farha2019ms}]
&17.2 & 63.1 & 56.2 &24.3 & 65.5 & 70.3 &47.4 & 73.1 & 73.5 \\
&ASformer  [\citenum{yi2021asformer}]
&14.1 & 60.8 & 55.3 &20.3 & 62.8 & 69.4 &44.8 & 72.9 & 71.9 \\
&G-TAD [\citenum{xu2020g}]
&23.0 & 66.1 & 64.0 &29.6 & 66.5 & 74.2 &53.3 & 76.0 & 77.9 \\
&ActionFormer [\citenum{zhang2022actionformer}]
&28.7 & {70.0} & {67.6} &35.5 & {70.9} & {76.4} &59.3 & {77.1} & {81.5} \\
&E2E-Spot [\citenum{hong2022spotting}]
&25.9 & 69.4 & 65.7 &33.8 & 70.4 & 75.4 &55.5 & 76.5 & 79.5 \\
\midrule
{\color{black}  I3D [\citenum{carreira2017quo}]} & {\color{black}E2E-Spot [\citenum{hong2022spotting}]} &{\color{black}22.7} & {\color{black}59.7} & {\color{black}68.7} & {\color{black}27.1} & {\color{black}60.7} & {\color{black}74.2} & {\color{black}51.9} & {\color{black}67.7} & {\color{black}78.3} \\
{\color{black}  VTN [\citenum{neimark2021video}]} & {\color{black}E2E-Spot [\citenum{hong2022spotting}]} &{\color{black}14.8} & {\color{black}58.3} & {\color{black}56.7} & {\color{black}20.0} & {\color{black}59.4} & {\color{black}68.2} & {\color{black}39.7} & {\color{black}63.1} & {\color{black}73.1}\\
\midrule
TSM [\citenum{lin2019tsm}]
&MS-TCN [\citenum{farha2019ms}]
& 21.7 & 67.3 & 58.6 & 30.4 & 69.5 & 73.0 &50.2 & 74.0 & 75.3 \\
&ASformer  [\citenum{yi2021asformer}]
& 17.6 & 61.9 & 57.5 & 25.5 & 64.0 & 74.2 &46.0 & 72.9 & 74.0\\
&G-TAD [\citenum{xu2020g}]
& 16.9 & 62.5 & 55.2 & 29.8 & 66.9 & 74.8 & 39.8 & 70.1 & 67.2 \\
&ActionFormer [\citenum{zhang2022actionformer}]
& 22.4 & 65.7 & 60.3 & 31.0 & 68.2 & 74.7 & 52.4 & 73.8 & 74.9 \\
&E2E-Spot [\citenum{hong2022spotting}]
& {31.4} & {71.4} & {68.7} & 39.5 & {72.3} & {77.9} & 60.6 & {78.4} & {82.1} \\
\midrule
TSM[\citenum{lin2019tsm}] &{$\fest{}$} &\textbf{40.3} & \textbf{75.2} & \textbf{74.0} & \textbf{48.0} & \textbf{76.5} & \textbf{82.4} &\textbf{68.4} & \textbf{80.0} & \textbf{87.2} \\
\bottomrule
\end{tabular}
}
\label{tab:results_full}
\end{table}
\normalsize

{\color{black} \section{Additional Ablation Studies} \label{ablation_sup}
In this section, we provide additional ablation studies as well as more implementation details.

\subsection{Frame-wise versus Clip-wise Feature Extractor}
We would like to clarify the advantages of frame-wise over clip-wise feature extraction.
\begin{itemize}[leftmargin=*]
    \item Temporal Precision: To use a clip-wise feature extractor, we can divide the input video into non-overlapping segments and extract one feature vector per segment, which is a common technique in many TAL and TAS tasks. To investigate this, we conducted an experiment where 96-frame video clips were divided into 6-frame segments, with features extracted using I3D [\citenum{carreira2017quo}] for each segment. The resulting 16 feature vectors were interpolated back to 96 frames using PyTorch’s F.interpolate function. As shown in Table~\ref{tab:ablation}(a), this approach produces temporally coarse features, leading to inadequate performance in precise event detection tasks.
    \item Efficiency and Scalability: An alternative approach is to stride a clip-wise feature extractor to obtain per-frame feature densely. However, this approach introduces significant computational overhead as each frame is processed multiple times in overlapping windows. This overhead makes end-to-end feature learning or fine-tuning impractical. In contrast, our frame-wise approach processes each frame only once, enabling the training of much longer sequences (hundreds of frames) in an end-to-end manner on a single GPU.
\end{itemize}

\subsection{Skeleton-based Feature Extractor}
We recognize the potential of using human pose estimation for representation learning and its ability to generalize to other domains. To explore this, we conducted additional experiments leveraging skeleton-based representations for \fff{} event detection in \dataset{}. We used MMPose\footnote{\url{https://mmpose.readthedocs.io/en/latest/}} to extract player key points from original 1280x720 resolution images, generating skeleton data. Two advanced skeleton feature extractors ST-GCN++ [\citenum{duan2022pyskl}] (GCN-based) and PoseConv3D [\citenum{duan2022revisiting}] (CNN-based) were evaluated. The extracted skeleton features were processed with the \fest{} head architecture for classification and localization. The results are summarized in Table~\ref{tab:skeleton-based}.

Key findings include:
\begin{itemize}[leftmargin=*]
    \item Among the two skeleton-based methods, ST-GCN++ demonstrated better overall performance.
    \item Visual features extracted from RGB images using TSM consistently outperformed skeleton-based methods in all three granularities. This is likely because many event types in \dataset{} include information such as shot direction and shot outcomes, which skeletal data cannot capture.
    \item Skeleton-based methods excel in computational efficiency and interpretability, requiring fewer parameters and offering faster inference, while directly highlighting player movements and poses.
\end{itemize}

While skeleton-based approaches may not fully match the performance of RGB-based models for \dataset{}, they offer unique benefits, particularly in terms of speed and transparency. We plan to further investigate skeleton-based methods and their integration with visual features in future work.

\subsection{Input image resolution}
We conducted additional experiments to analyze the effects of using different resolution inputs on model performance. The results are summarized in Table~\ref{tab:resolution}.

First, we evaluated \dataset{} using TSM as the video encoder with input resolutions of 224$\times$224, 336$\times$336, and 448$\times$448. The results show a consistent improvement in performance as resolution increases, though the gains diminish at higher resolutions (e.g., 448$\times$448). This suggests that while higher resolutions can provide additional visual details, the marginal benefits decrease beyond a certain point.

We also tested \fest{} with SlowFast as an encoder on higher-resolution inputs (336$\times$336). We experimented with two variants: SlowFast 4$\times$16 and SlowFast 8$\times$8. Despite higher complexity, SlowFast 4$\times$16 underperformed TSM, likely due to its lower temporal resolution, which limits its ability to capture subtle differences. SlowFast 8$\times$8 achieved slightly better performance than SlowFast 4$\times$16 and marginally outperformed TSM in Edit score (+0.2) but lagged in F$1_{evt}$ and F$1_{elm}$ metrics.

The 224$\times$224 resolution remains a common choice in video analytics due to its efficiency and compatibility with pre-trained models. Balancing complexity, performance, and efficiency, we selected 224$\times$224 and TSM as the default configuration for \fest{}. We will include the above analysis in the revised manuscript to clarify the trade-offs between resolution, complexity, and performance.

\subsection{Choice of CTX module}
For the CTX module, we acknowledge that transformer-based models have demonstrated superior efficiency in modeling long-range dependencies. To ensure our choice was justified, we conducted comparative experiments using a Bidirectional GRU (BiGRU) and a Transformer Encoder for the CTX stage. The results are summarized in Table~\ref{tab:ablation}(f). As the results indicate, the performance of the two module choices is comparable, with BiGRU slightly outperforming the Transformer Encoder in our \fest{} system. We attribute this to the relatively short event sequences passed to the CTX module, which typically contains fewer than 20 events per 96-frame input clip. Under these conditions, the BiGRU effectively models the necessary temporal context with fewer parameters and lower computational overhead compared to the Transformer Encoder.}

\begin{table}
\caption{Skeleton-based method compared with TSM + \fest{}. ``Params(M)'' refers to the number of model parameters, and ``Inference time (ms)'' refers to the per-frame inference time on a Nvidia RTX 4090 GPU.}
\vspace{-1mm}
\centering
\small
\resizebox{0.95\textwidth}{!}{
\begin{tabular}{@{}lcccccccccccc@{}}
\toprule
&  & & \multicolumn{3}{c}{$\dataset{}\,(G_{high})$} & \multicolumn{3}{c}{$\dataset{}\,(G_{mid})$} & \multicolumn{3}{c}{$\dataset{}\,(G_{low})$} \\
\cmidrule(r){4-6} \cmidrule(lr){7-9} \cmidrule(l){10-12} 
{Experiment} & Params (M) & Inference time (ms)  & F$1_{evt}$ & F$1_{elm}$ & Edit & F$1_{evt}$ & F$1_{elm}$ & Edit & F$1_{evt}$ & F$1_{elm}$ & Edit \\
\midrule
{TSM + \fest{}} & 5.6 & 10.6 &40.3 & 75.2 & 74.0 & 48.0 & 76.5 & 82.4 &68.4 & 80.0 & 87.2 \\
\midrule
{\color{black}  ST-GCN++ [\citenum{duan2022pyskl}]} & 2.3 & 4.0 &{\color{black}25.4} & {\color{black}62.1} & {\color{black}56.1} & {\color{black}32.4} & {\color{black}63.9} & {\color{black}63.5} & {\color{black}55.1} & {\color{black}69.4} & {\color{black}73.2} \\
{\color{black}  PoseConv3D [\citenum{duan2022revisiting}]} & 6.8 & 6.4 &{\color{black}20.1} & {\color{black}54.5} & {\color{black}53.2} & {\color{black}26.0} & {\color{black}55.4} & {\color{black}61.9} & {\color{black}48.8} & {\color{black}63.0} & {\color{black}69.7} \\
\bottomrule
\end{tabular}
}
% \vspace{-10px}
\label{tab:skeleton-based}
\end{table}
\normalsize

\begin{table}
\caption{Ablation study on input image resolution. ``Params(M)'' refers to the number of model parameters. ``FLOPs'' refers to floating-point operations per second.}
\vspace{-1mm}
\centering
\small
\begin{tabular}{@{}lcccccccccccc@{}}
\toprule
& & & & & \multicolumn{3}{c}{$\dataset{}\,(G_{high})$}  \\
\cmidrule(r){6-8}
{Video encoder} &Head arch. &Resolution & Params (M) & FLOPs  & F$1_{evt}$ & F$1_{elm}$ & Edit  \\
\midrule
TSM & \fest{} & 224$\times$224 & 5.6 & 77.3 &40.3 & 75.2 & 74.0  \\
\midrule
TSM & \fest{} & 336$\times$336 & 5.6 & 77.3 &43.2 &77.1 &74.8 \\
TSM & \fest{} & 448$\times$448 & 5.6 & 77.3 &44.4 & 78.1 & 74.5\\
SlowFast 4$\times$16 & \fest{} & 336$\times$336 & 52.7 & 494.2 &37.4 & 73.7 &71.3 \\
SlowFast 8$\times$8 & \fest{} & 336$\times$336 & 52.8 & 903.7 &41.0 &74.6 &75.0 \\
\bottomrule
\end{tabular}
\label{tab:resolution}
\end{table}
\normalsize

\section{Examples of Error Event Sequences} \label{error_sequences}
In this section, we discuss examples of predicted event sequences alongside their corresponding ground truth sequences. These examples illustrate instances of logical errors or uncommon practices observed in predictions generated by TSM + E2E-Spot.

\textbf{Example 1 (Logical Error):}
\begin{small}
\begin{verbatim}
Clip: 20190308-W-Indian_Wells-R64-Serena_Williams-Victoria_Azarenka_173593_173770
Far-end player: Victoria Azarenka (right-handed)
Near-end player: Serena Williams (right-handed)

Predicted sequence:
near_deuce_serve_-_-_W_-_in -> 
far_deuce_return_fh_gs_CC_-_in -> 
near_deuce_stroke_fh_gs_DL_-_winner -> 
far_ad_stroke_bh_slice_DL_-_forced-err

Ground truth sequence:
near_deuce_serve_-_-_W_-_in -> 
far_deuce_return_fh_gs_CC_-_in -> 
near_deuce_stroke_fh_gs_DL_-_in -> 
far_ad_stroke_bh_slice_DL_-_forced-err
\end{verbatim}
\end{small}
In this rally between Victoria Azarenka and Serena Williams, the predicted event sequence contains a clear logical error: the sequence labels a shot as a ``winner'' but is followed by another shot labeled as ``forced-err'', which contradicts the definition of a ``winner''.

\textbf{Example 2 (Logical Error):}
\begin{small}
\begin{verbatim}
Clip: 20190308-W-Indian_Wells-R64-Serena_Williams-Victoria_Azarenka_64883_65114
Far-end player: Serena Williams (right-handed)
Near-end player: Victoria Azarenka (right-handed)

Predicted sequence:
near_deuce_serve_-_-_T_-_in -> 
far_middle_return_bh_gs_DM_-_in -> 
near_middle_stroke_fh_gs_CC_-_in -> 
far_deuce_stroke_fh_gs_DM_-_in -> 
near_deuce_stroke_fh_gs_DL_-_winner

Ground truth sequence:
near_deuce_serve_-_-_T_-_in -> 
far_middle_return_bh_gs_DM_-_in -> 
near_middle_stroke_fh_gs_CC_-_in -> 
far_deuce_stroke_fh_gs_DM_-_in -> 
near_middle_stroke_fh_gs_IO_-_winner
\end{verbatim}
\end{small}
The error in this prediction occurs in the final event, where the predicted ``near\_deuce\_stroke\_fh\_gs\_DL...'' contradicts the ground truth ``near\_middle\_stroke\_fh\_gs\_IO...''. The prediction does not logically follow from the previous event where the far-end player directed the ball down the middle.

\textbf{Example 3 (Uncommon Practice):}
\begin{small}
\begin{verbatim}
Clip: 20130607-M-Roland_Garros-SF-Novak_Djokovic-Rafael_Nadal_108769_108956
Far-end player: Rafael Nadal (left-handed)
Near-end player: Novak Djokovic (right-handed)

Predicted sequence:
near_deuce_serve_-_-_T_-_in -> 
far_middle_return_bh_gs_CC_-_in-> 
near_ad_stroke_bh_gs_CC_-_winner

Ground truth sequence:
near_deuce_serve_-_-_T_-_in -> 
far_middle_return_fh_gs_CC_-_in-> 
near_ad_stroke_bh_gs_CC_-_winner
\end{verbatim}
\end{small}
In this rally between Novak Djokovic and Rafael Nadal, the predicted sequence suggests an uncommon practice: Nadal, a left-handed player, is unlikely to return a deuce court serve to T using his backhand. Typically, a left-hander would use a forehand for such a shot, indicating a likely error in the predicted event.

{\color{black}\section{Impact of the Event Localizer to the Whole \fest{} System} \label{impact_lcl}
To understand the impact of the Event Localizer (LCL) on the performance of the overall \fest{} system, we conducted additional analysis and included an ``F$1_{lcl}$'' column in Table~\ref{tab:impact_lcl}, which evaluates the precision of the LCL module in identifying event moments with tight temporal tolerance. The table compares the F$1_{lcl}$ metric with the overall system metrics (F$1_{evt}$ and Edit) across various datasets. All use TSM as video encoder and \fest{} as the head architecture.

We observe that the performance of \fest{} is positively correlated with the quality of the LCL module. For example, datasets like FineDiving and ShuttleSet, which have high-performing LCL modules, result in superior downstream performance (F$1_{evt}$ and Edit). Conversely, on datasets like CCTV-Pipe, where the LCL module performs less effectively, \fest{}’s overall performance is less ideal.

However, it is important to highlight that even when the LCL module does not perform well, our method still outperforms other state-of-the-art methods (as shown in Table~\ref{tab:general} in the paper). Therefore, a very good-performing LCL module is not a hard prerequisite.
}

\begin{table}
\caption{Ablation study on input image resolution.}
\vspace{-1mm}
\centering
\small
\begin{tabular}{@{}lcccccccccccc@{}}
\toprule
Dataset & F$1_{lcl}$ & F$1_{evt}$ & Edit  \\
\midrule
\dataset{} ($G_{high}$) & 86.7 & 40.3 & 74.0\\
ShuttleSet & 97.9 & 70.7 & 77.1\\
FineDiving & 94.2 & 77.6 & 95.1\\
FineGym & 84.0 & 70.9 & 70.7\\
CCTV-Pipe & 71.9 & 37.0 & 39.5\\
\bottomrule
\end{tabular}
\label{tab:impact_lcl}
\end{table}
\normalsize

\section{Limitations and Social Impact} \label{limitation}
In addressing the limitations of our current work, we acknowledge that the expansion of our dataset to include a broader range of videos from both professional and lower-tier matches is an essential yet exceedingly time-consuming and labor-intensive task. The enhancement of our dataset is imperative for providing a more comprehensive analysis that spans various levels of play.

Furthermore, the primary objective of this project is to extend the scope of our tennis analytics from exclusively focusing on elite professional athletes to encompassing a wider audience. This includes semi-professional players, collegiate athletes, junior competitors, and general tennis enthusiasts. By broadening our analytical reach, we aim to democratize access to advanced sports analytics, enabling players at all levels to refine their techniques and strategies.

Socially, the implications of our work are significant. By making sophisticated analytics available to a more diverse group of users, we can contribute positively to public health and fitness. Access to detailed performance data allows individuals to make informed decisions about their training regimes, thus enhancing their overall sports skills and encouraging a healthier lifestyle. This democratization not only fosters a greater appreciation and understanding of tennis but also motivates a broader spectrum of the population to engage actively in sports, thereby promoting physical well-being and health consciousness across communities.

\end{document}